\documentclass{article}

    \usepackage[final]{template/neurips_2024}

\usepackage[utf8]{inputenc} 
\usepackage[T1]{fontenc}    
\usepackage{hyperref}       
\usepackage{url}            
\usepackage{booktabs}       
\usepackage{amsfonts}       
\usepackage{nicefrac}       
\usepackage{microtype}      
\usepackage{xcolor}         
\usepackage{subfigure}

\usepackage{amsmath,amsfonts,bm}
\usepackage{mathtools}

\def\eqref#1{equation~\ref{#1}}

\def\1{\bm{1}}

\def\ve{{\bm{e}}}

\def\vg{{\bm{g}}}

\def\vw{{\bm{w}}}

\def\mH{{\bm{H}}}

\def\mW{{\bm{W}}}

\DeclareMathAlphabet{\mathsfit}{\encodingdefault}{\sfdefault}{m}{sl}
\SetMathAlphabet{\mathsfit}{bold}{\encodingdefault}{\sfdefault}{bx}{n}

\def\gS{{\mathcal{S}}}

\def\sR{{\mathbb{R}}}

\newcommand{\norm}[1]{\left\lVert#1\right\rVert}

\newcommand{\transpose}{^\top}

\newcommand{\nll}{\mathrm{NLL}}
\newcommand{\snr}{\mathrm{SNR}}
\newcommand{\sqnr}{\mathrm{SQNR}}

\newenvironment{ls}{
\begin{enumerate}
  \vspace{-3mm}
  \setlength{\leftskip}{-3mm}
  \setlength{\itemsep}{0pt}
  \setlength{\parskip}{0pt}
  \setlength{\parsep}{0pt}
}{\end{enumerate}}

\makeatletter
\renewcommand{\paragraph}{%
  \@startsection{paragraph}{4}%
  {\z@}{0.25ex \@plus 0.25ex \@minus .5ex}{-1em}%
  {\normalfont\normalsize\bfseries}%
}
\makeatother

\title{Scaling Laws for Post Training Quantized Large Language Models}

\author{%
  Zifei Xu \thanks{Equal contributions.} \\
  d-Matrix\\
  Santa Clara, CA, USA\\
  \texttt{xuzifei@d-matrix.ai} \\
  \And
  Alexander Lan \footnotemark[1] \space \thanks{Work done when the author was an intern at d-Matrix.}  \\
  Yale University\\
  New Haven, CT, USA \\
  \texttt{alex.lan@yale.edu} \\
  \And
  Wanzin Yazar \\
  d-Matrix\\
  Santa Clara, CA, USA\\
  \texttt{wyazar@d-matrix.ai} \\
  \And
  Tristan Webb \\
  d-Matrix\\
  Santa Clara, CA, USA\\
  \texttt{twebb@d-matrix.ai} \\
  \And
  Sayeh Sharify \\
  d-Matrix\\
  Santa Clara, CA, USA\\
  \texttt{sayehs@d-matrix.ai} \\
  \And
  Xin Wang \\
  d-Matrix\\
  Santa Clara, CA, USA\\
  \texttt{xwang@d-matrix.ai} \\
}

\begin{document}

\maketitle

\begin{abstract}

Generalization abilities of well-trained large language models (LLMs) are known to scale predictably as a function of model size.  
In contrast to the existence of practical scaling laws governing pre-training, the quality of LLMs after post-training compression remains highly unpredictable, often requiring case-by-case validation in practice. 
In this work, we attempted to close this gap for post-training weight quantization of LLMs by conducting a systematic empirical study on multiple LLM families quantized to numerous low-precision tensor data types using popular weight quantization techniques.  
We identified key scaling factors pertaining to characteristics of the local loss landscape, based on which the performance of quantized LLMs can be reasonably well predicted by a statistical model. 

\end{abstract}
\section{Introduction}

Large language models (LLMs) based on the transformer architecture~\citep{vaswani2023attention} are known to obey empirical scaling laws. 
An LLM's generalization abilities, measured by the negative-log-likelihood (NLL) loss in next-token prediction, are predictably related to increases in parameter count, pre-training data volume, and computation cost~\citep{kaplan2020scaling,dettmers2023case,henighan2020scaling,alabdulmohsin2022revisiting,su2024unraveling,song2024resource,muennighoff2023scaling,bordelon2024dynamical,bahri2024explaining}.

Thanks to the guidance from these scaling laws, pre-training of LLMs, a notoriously expensive computation in practice, enjoys a certain degree of confidence in return on investment.   
However, for these LLMs to run efficiently on a target accelerator for inference, they often need to undergo post-training compression, such as quantization~\citep{gholami2021survey,frantar2022gptq,park2024lutgemm,kim2023finequant,kim2024squeezellm,yao2022zeroquant}.  

Post-training quantization (PTQ) is a process that attempts to preserve a trained LLM's generalizability, while performing its computation with low-precision data types. 
Because PTQ involves many additional factors, it introduces significant  uncertainty into the quality of the final model, in many cases completely obscuring the predictability prescribed by the pre-training scaling laws.  
This makes PTQ a business of trial-and-error~\citep{huang2024good,sharify2024combining,yuan2023benchmarking,hu2022empirical}, lacking the useful practical guidance from scaling laws like those that govern pre-training. 

Beyond just algorithmic complexity, PTQ also becomes incredibly time and compute intensive when one attempts to find the optimal quantization format and model parameter count given fixed memory, compute, and data format constraints. A simple illustration of this trade-off is the comparison between a larger model quantized to a lower bit format and a smaller model quantized to a higher bit format, a search space that requires a lot of iterations, and consequently, significant time and compute.

In this work, we attempted to close this gap in knowledge by systematically studying the empirical scaling of extra factors involved in PTQ in addition to the pre-trained NLL loss.   
We briefly enumerate below all factors considered.  

\begin{ls}
  \item \textbf{Loss of pre-trained LLM.}  A known scaling law governs the relationship between LLM training parameters and the quality of the resulting model. Intuitively, a better trained model would also have better performance in a quantized state, so the initial loss of a pre-trained LLM is highly relevant to profiling the quantized loss landscape. Section~\ref{sec:nll_loss} is dedicated it.
  \item \textbf{Local loss landscape of pre-trained LLM.}  Because quantization is a specific perturbation to the trained network, the resulting loss due to the perturbation depends not only on the converged NLL loss, but also on how steeply the loss changes in the neighborhood of convergence ~\citep{frumkin2023jumping,nahshan2020loss,evci2020difficulty}.  Section~\ref{sec:loss_landscape} is dedicated to understanding how the local loss landscape changes with scale.
  \item \textbf{Low-precision data type for quantization.}  Numerous novel tensor data types for efficient inference have emerged recently~\citep{microscaling2023,dettmers2023qlora,agrawal2024exmy,guo2022ant}. Intuitively, both the tensor data type and its numerical precision would correlate with the quality of quantization, and Section~\ref{sec:data_type} is dedicated to its scaling. 
  \item \textbf{PTQ algorithm.}  After aggressive low-precision quantization, certain PTQ optimization algorithms are commonly used to recover some model quality~\citep{frantar2022gptq,xiao2024smoothquant,lin2024awq,lee2024owq}.  These methods typically minimizes local quantization error as opposed to direct global loss optimization as in quantization-aware fine-tuning (\emph{e.g.} \citealt{li2023loftq,jeon2024l4q}).  Section~\ref{sec:gptq} is dedicated to profiling how those properties scale. 
\end{ls}

We show with concrete examples (for procedural details see Section~\ref{sec:methods}), that all the above factors have underlying empirical scaling laws for certain LLM families.  
Incorporating these empirical rules, in Section~\ref{sec:prediction}, we build a predictive statistical model that takes the above factors as input and predicts the outcome of a PTQ procedure on unseen LLMs at a reasonable accuracy.

\section{Factors subject to scaling for LLM PTQ}

\begin{figure}[t!]
  \centering
  \subfigure{
    \includegraphics[width=0.48\textwidth]{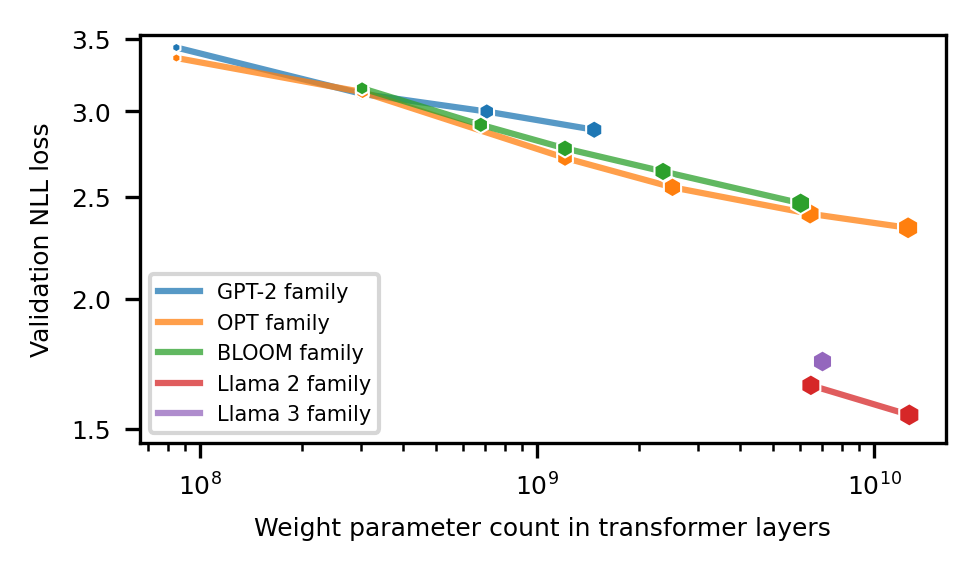}
  }
  \subfigure{
    \includegraphics[width=0.48\textwidth]{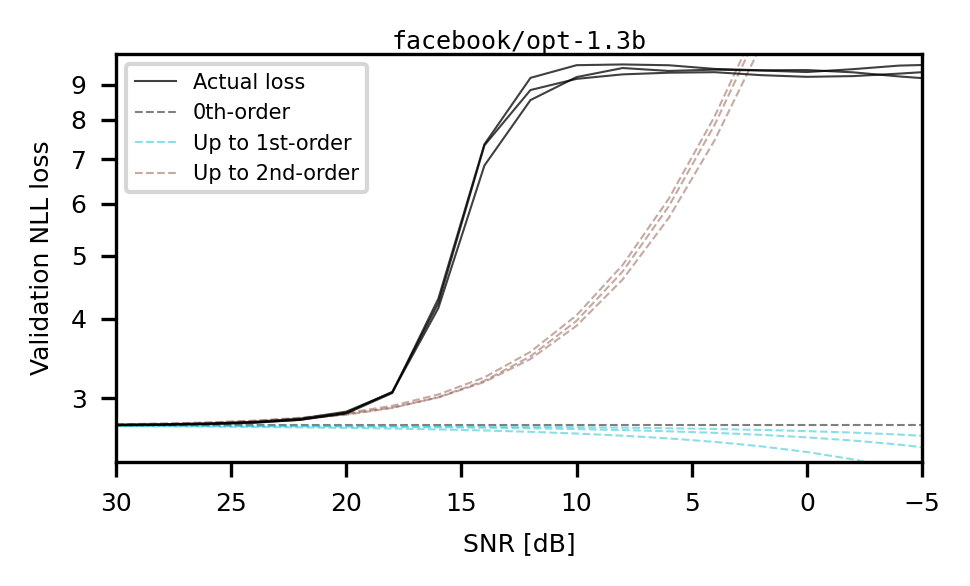}
  }
  \caption{
    Left: \textbf{Scaling of pre-trained NLL loss.}
    NLL losses evaluated on the validation split of the WikiText-2 dataset are plotted against the total parameter counts in the transformer layers' weight tensors. 
    Model families are color-coded and the symbol sizes encode the weight parameter count, a convention shared by following figures.
    Right: \textbf{Local radial loss landscape mapping.}  
    Shown here is measurement of the \emph{typical} loss landscape in the neighborhood of pre-trained weights, by evaluation of the loss along typical radial perturbations, 3 independent instances illustrated for \texttt{opt-1.3b}, together with their Taylor series approximations.
  }
  \label{fig:loss_scaling_and_landscape}
\end{figure}

\begin{figure}[h!]
  \centering
  \subfigure{
    \includegraphics[width=0.48\textwidth]{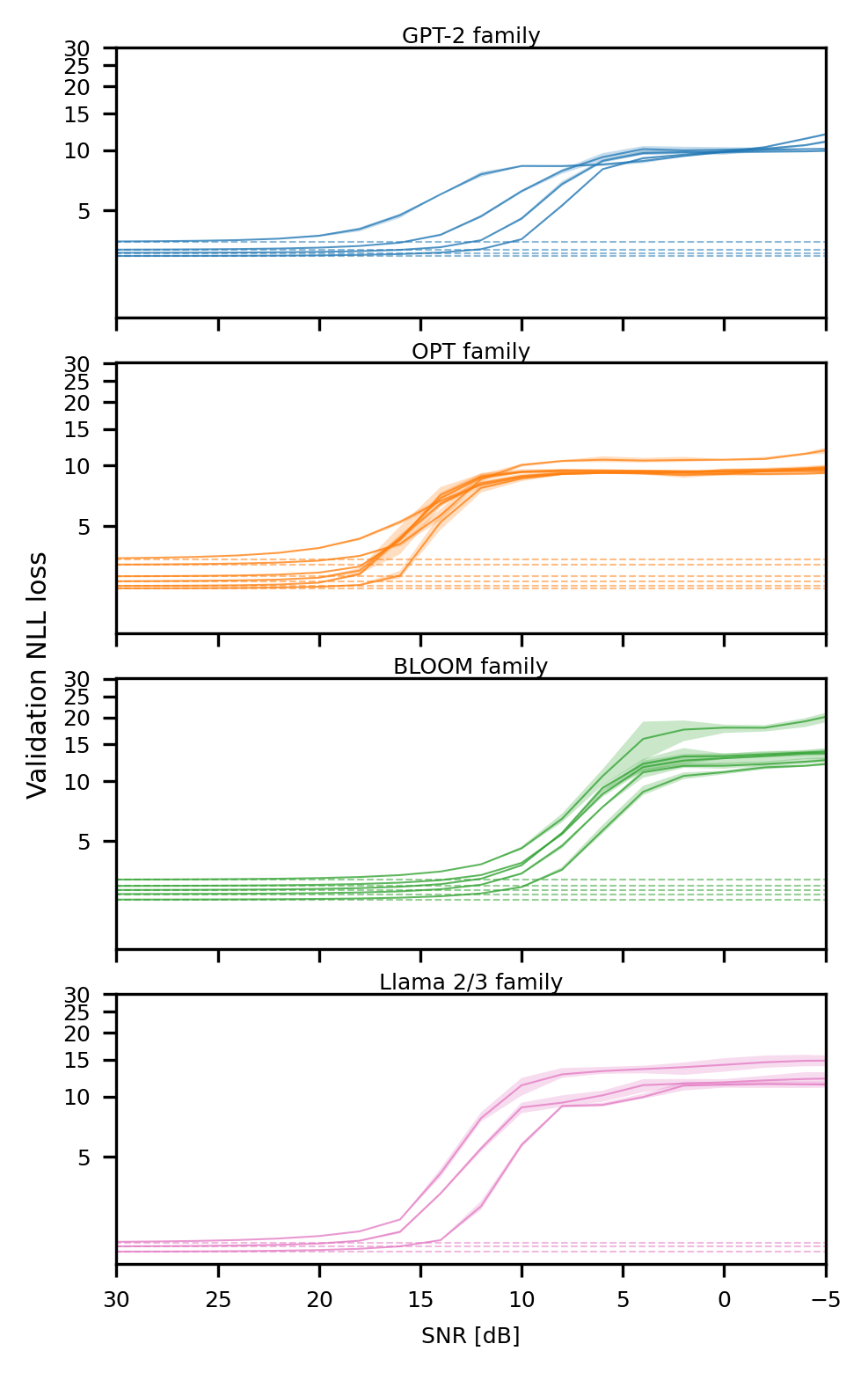}
  }
  \subfigure{
    \includegraphics[width=0.48\textwidth]{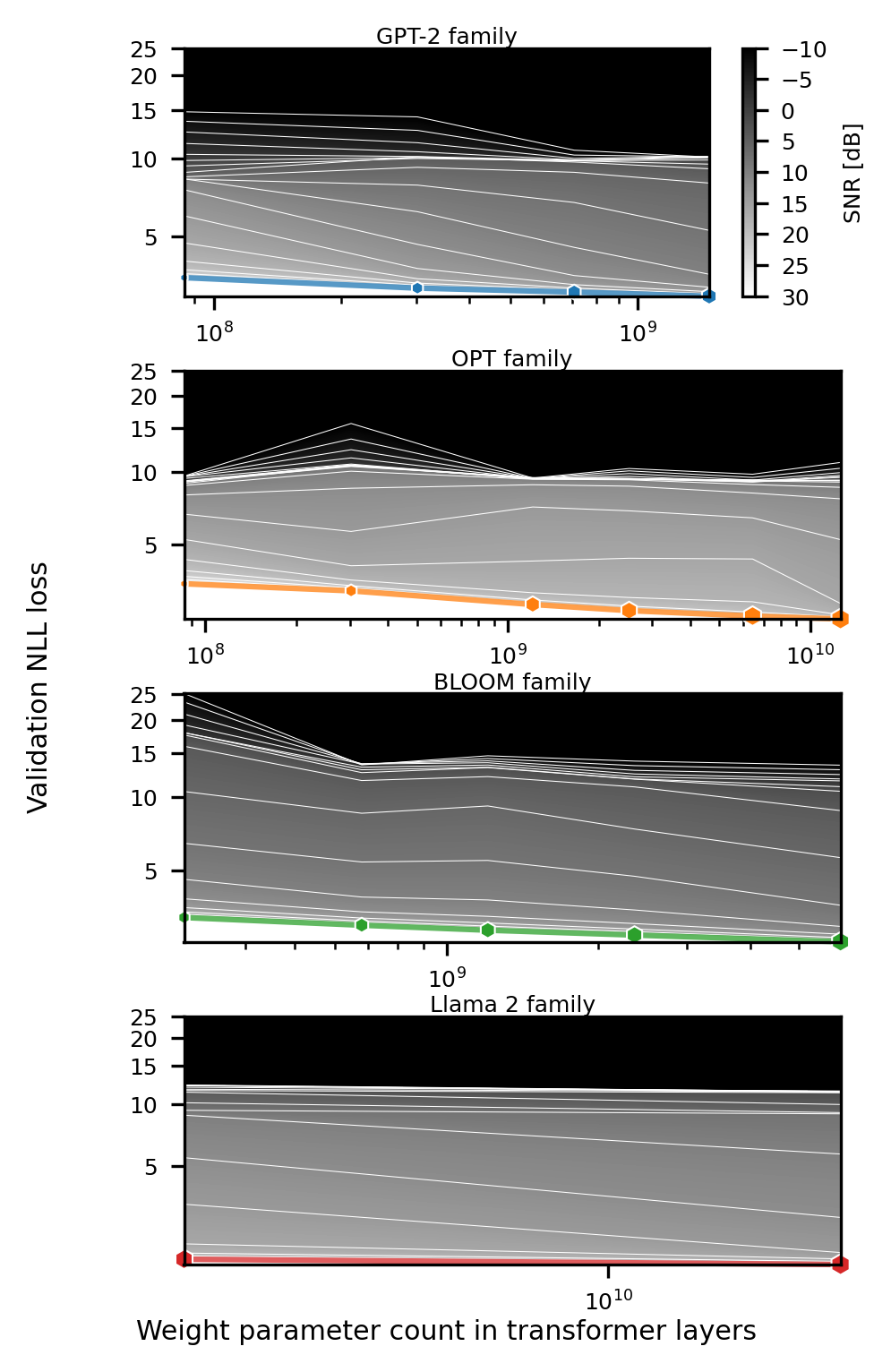}
  }
  \caption{
    Left: \textbf{Local loss landscape of LLMs grouped in families.}
    Shown are the mean (colored curves) and range (colored shades) of 3 independent measurements for each model. 
    The typical characteristics are common to all models. 
    Within a family, larger models tend to have flatter local loss landscape, in a predictable manner. Right: \textbf{Scaling of local loss landscape as a function of LLM size.} 
    We plot NLL loss against weight parameter count, with typical perturbation SNR as a gray-scale heat map.  
    Thin white iso-SNR curves are at 2 dB increments.  
    With OPT family as the only exception, vertical spacing of these iso-SNR curves is shorter in large models than in small ones of the same family, suggesting flatter local minima at larger model sizes.   
  }
  \label{fig:lls_family_and_heatmap}
\end{figure}

\subsection{Loss of pre-trained LLM}
\label{sec:nll_loss}

First, we recapitulate one of the original scaling laws on well trained LLMs with no data limit~\citep{kaplan2020scaling}. 
We visualize in Figure~\ref{fig:loss_scaling_and_landscape} (left) this scaling law with our experiments (see Section~\ref{sec:methods} for details).  
The GPT-2, OPT and BLOOM model families roughly follow one power law, whereas models in the Llama 2/3 family track a different, but qualitatively similar path.

\subsection{Characteristics of local loss landscape}
\label{sec:loss_landscape}

Next, we characterize another crucial factor intrinsic to the LLM itself, its local loss landscape.

A quantization of network weight $\vw$~\footnote{Here we denote by vector $\vw$ a flattened version of all weight matrices $\left(\mW_1, \cdots, \mW_L\right)$ of the network that are subject to quantization.} can be considered as a perturbation $\vw \rightarrow \vw + \Delta\vw= Q(\vw)$, where $Q$ is a quantizer, and the resulting loss of the quantized network becomes $\nll(\vw + \Delta\vw)$ from the pre-trained $\nll(\vw)$.  
The resulting loss is a function not only of the pre-trained weight $\vw$, but also of the perturbation $\Delta\vw$, often approximated by Taylor expansion,  
\begin{align*}
    \nll(\vw + \Delta\vw) &= \nll(\vw) \\
    &+ \vg \transpose \Delta\vw + \frac 1 2 \Delta\vw \transpose \mH \Delta\vw \\
    &+ O(\norm{\Delta\vw}^2) .
\end{align*}
Here $\vg$ and $\mH$ are the gradient and Hessian at $\vw$, and $\norm{\cdot}$ is the $\ell_2$-norm.

As the absolute magnitude of $\vw$ scales with dimensionality (see Appendix~\ref{sec:weight_norm}), we use signal-to-noise ratio (SNR), a relative quantity to measure the magnitude of its perturbation $\Delta\vw$, 
\begin{align*}
    \snr(\vw, \Delta\vw) &= 20 \log_{10} \frac {\norm{\vw}} {\norm{\Delta\vw}} ,
\end{align*}
in decibel (dB).  
A higher SNR represents a smaller deviation $\Delta\vw$ from $\vw$. 
When the perturbation is due to quantization, \emph{i.e.}~$\Delta\vw = Q(\vw) - \vw$, SNR becomes signal-to-quantization-noise ratio (SQNR), 
\begin{align*}
    \sqnr(\vw) &= 20 \log_{10} \frac {\norm{\vw}} {\norm{Q(\vw) - \vw}} .
\end{align*}

Intuitively, the flatter the local loss landscape is near $\vw$, the less impact a same perturbation $\Delta\vw$ is to exert on the loss.
In Figure~\ref{fig:loss_scaling_and_landscape} (right), we show with an example LLM, the \emph{typical} local loss landscape in the neighborhood of pre-trained weights.  
We randomly sample a unit vector $\hat\ve \sim \gS^D$ from the $D$-dimensional unit sphere, $D$ being the dimensionality of $\vw$, and measure $\nll(\vw + \lambda \hat\ve)$ while sweeping $\lambda \in \sR^+$.  
We see that the typical radial loss is very \emph{step-like}: it stays relatively low and flat near $\vw$, then rises rapidly (faster than quadratic), and finally plateaus further away from $\vw$. 
These qualitative characteristics are shared by all LLMs of various sizes and from various families (Figure~\ref{fig:lls_family_and_heatmap}, left).  

We also find that, within the same LLM family, larger models have flatter local loss landscape than smaller ones, in a systematic way (Figures~\ref{fig:lls_family_and_heatmap}) for each family. 

\subsection{Low-precision data type for quantization}
\label{sec:data_type}

Now, we identify an extrinsic factor in PTQ process: the low-precision tensor data type for quantization.   
Note that we consider tensorial data types, not simply scalar numerical formats. 
In addition to traditional integer quantization that requires calibration, emerging standards such as microscaling (MX, \citealt{microscaling2023}) adopt more effective and efficient tensor data types, which we study in this work.
we also present a comparative study of traditional integer quantization in Appendix~\ref{sec:int_quantization}.  

We first ask how the magnitude of quantization errors $\Delta\vw = Q(\vw) - \vw$ vary across LLMs for certain data types.  
Despite the existence of significant scaling of $\norm{\vw}$ (see Appendix~\ref{sec:weight_norm} for further details), the SQNRs are relatively invariant across model families and model sizes, and vary across numerical data types in a highly predictable manner (see details in Appendix ~\ref{sec:data_format}).  
In contrast, NLL losses show a much more nonlinear and less predictable pattern, with a rough trend of lower precision data formats leading to higher losses (see details in Appendix ~\ref{sec:data_format}).

However, with certain choices of weight data type, the perturbation due to quantization is significantly flatter than the \emph{typical} flatness of the local loss landscape, which we shall elaborate in the next section.  

\begin{figure}[h]
    \centering
    \subfigure{
        \includegraphics[width=0.48\textwidth]{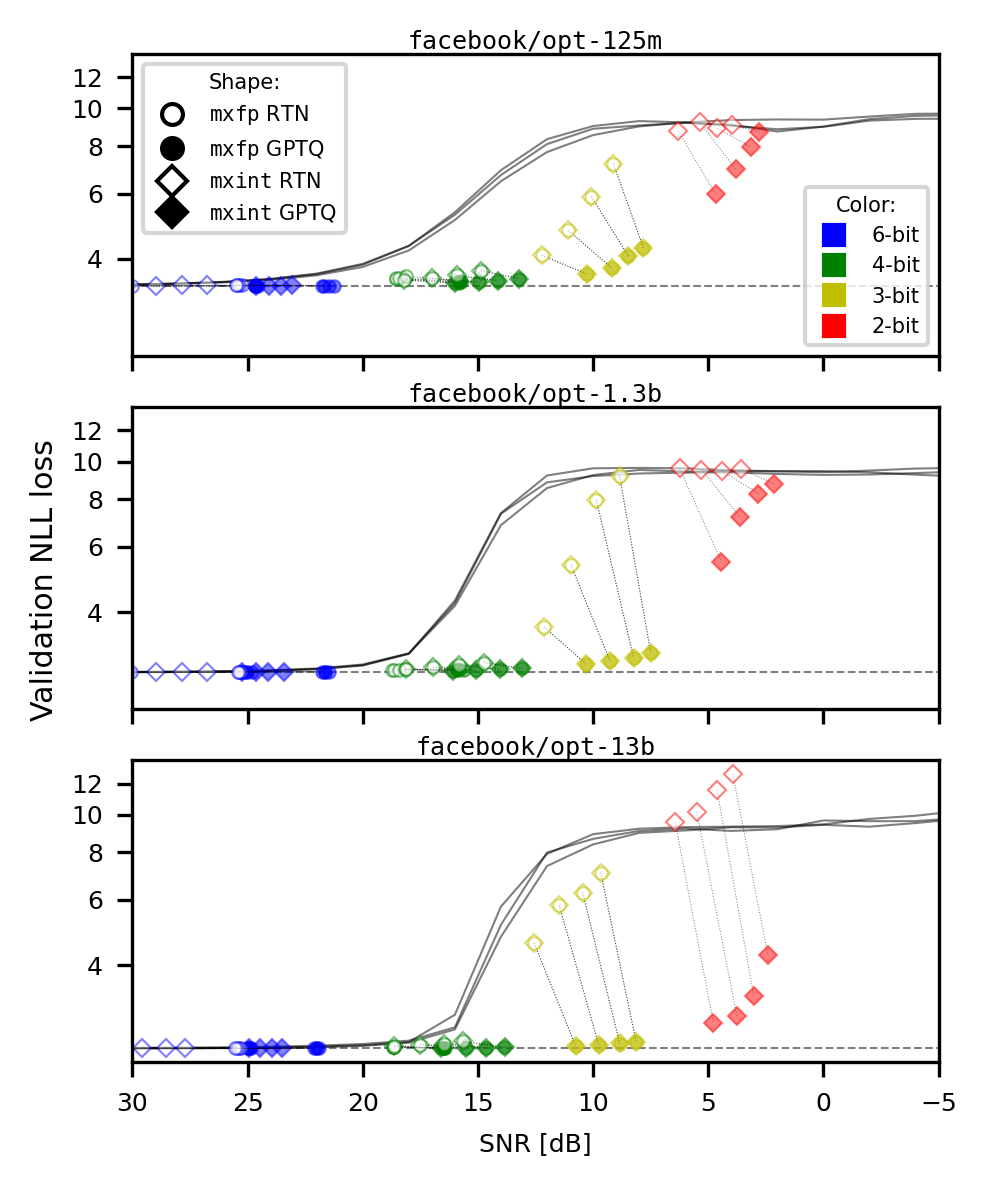}
        \label{fig:scaling_sqnr_nll_with_loss_landscape}
        }
    \hfill
    \subfigure{
        \includegraphics[width=0.48\textwidth]{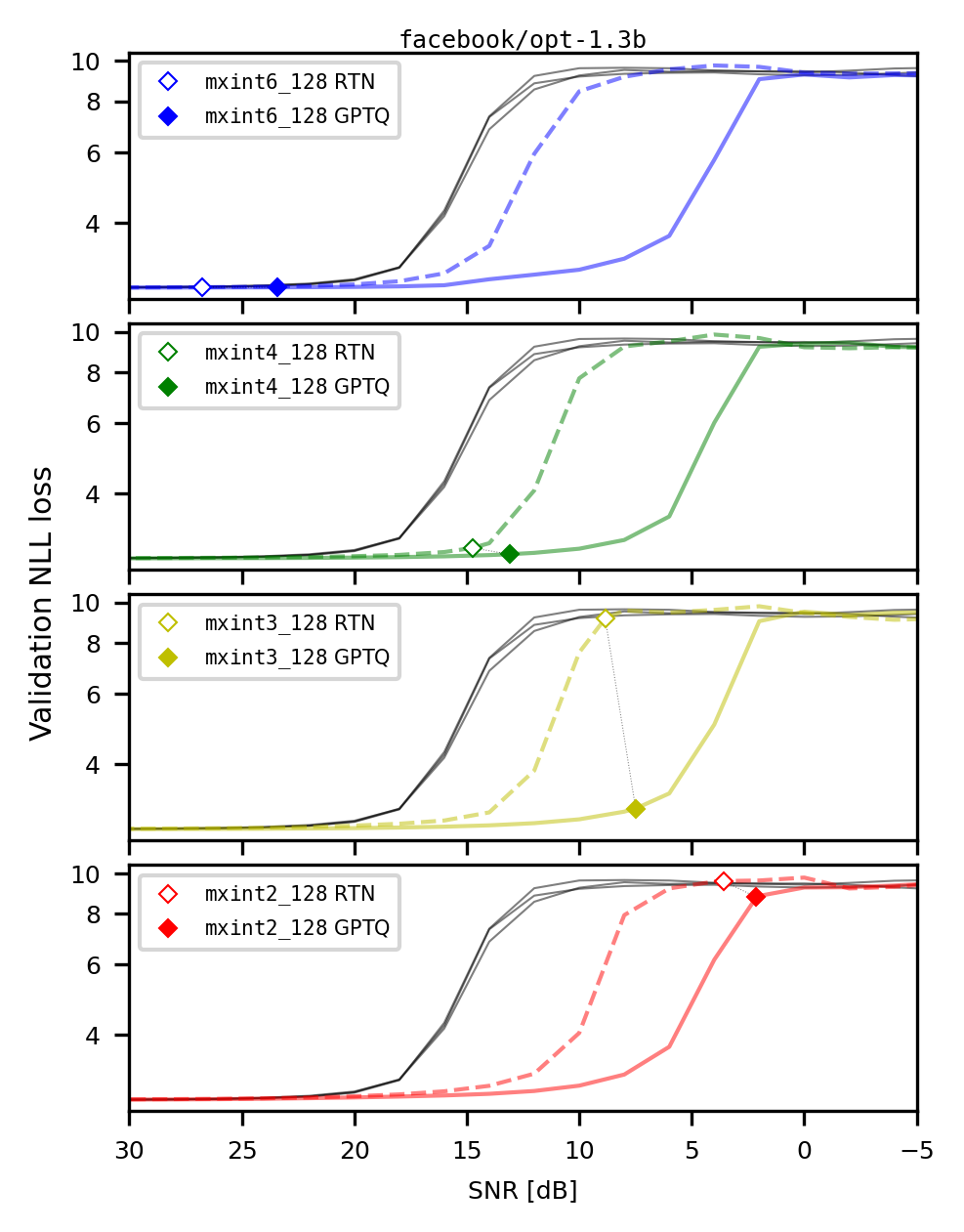}
        \label{fig:gptq_loss_landscape_explanation}
        }
    \caption{
    Left: \textbf{Scaling of SQNRs and NLL losses before and after PTQ, relative to the typical loss landscape.}  
    We show data from 3 members of the OPT model family, whose parameter counts are separated by 1 order of magnitude.  
    RTN (before PTQ, hollow symbols) and GPTQ (after PTQ, filled symbols) are plotted together with the typical radial loss landscape empirically mapped. 
    Right: \textbf{Local loss landscape underlying varied effectiveness of GPTQ acting on the same model quantized at different weight precision.}  
    Shown here are data of \texttt{opt-1.3b} quantized to \texttt{mxint6\_128}, \texttt{mxint4\_128}, \texttt{mxint3\_128} and \texttt{mxint2\_128}.  
    The colored, hollow or filled diamonds represent the SQNRs and NLL losses before and after GPTQ, respectively. 
    We further map the underlying radial loss landscape in the directions of typical random perturbation (thin gray lines), of RTN quantization (colored dashed lines) and of GPTQ quantization (colored solid lines).  
    }
    \label{fig:loss_landscape}
\end{figure}

\begin{figure}[h]
  \centering
  \includegraphics[scale=0.9]{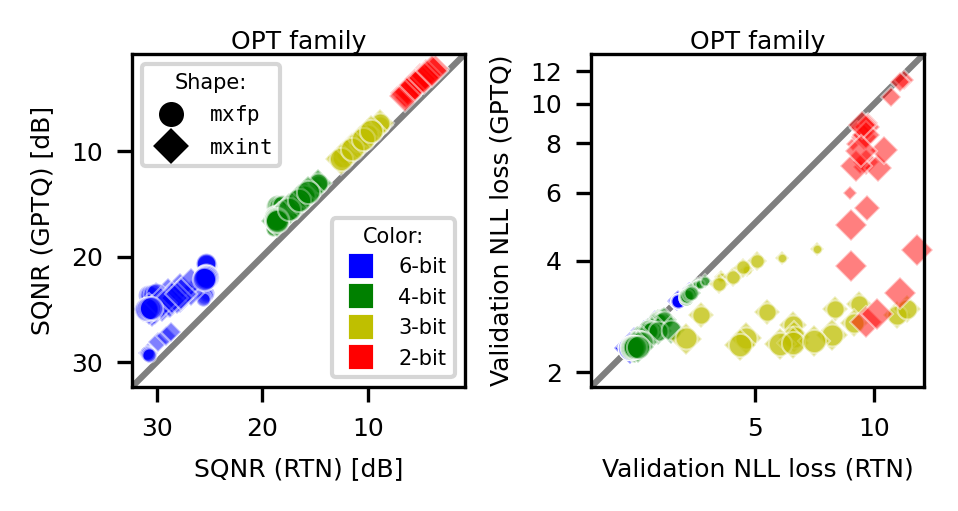} 
  \caption {
    \textbf{Changes in SQNRs and NLL losses resulting from GPTQ for OPT family.} 
    Numerical precision is color-coded and model size encoded by symbol size.  
    Diagonal line represents identity.  
  }
  \label{fig:direct_comparison_sqnr_nll_opt}
\end{figure}

\subsection{PTQ optimization method}
\label{sec:gptq}

Finally, we study another important extrinsic factor that contributes to the quality of quantized LLMs for inference, the PTQ optimization algorithm.  

To each model and for each weight data type, we applied an improved GPTQ procedure (see Section~\ref{sec:meth_ptq} for details) to further optimize the RTN quantized network. 
Figure~\ref{fig:loss_landscape} (left) shows 3 members of varied sizes from the OPT family.   
Apparently, the application of GPTQ generally reduced both the SQNR and NLL loss of the RTN model.  
The reduction in SQNR is relatively consistent across model sizes and data formats, whereas the reduction in NLL loss is highly variable as a function of model size and quantization precision in, however, a rather systematic way. 
An aggregation of direct comparisons of SQNRs and NLL losses before and after the GPTQ procedure for the OPT model family is presented in Figure~\ref{fig:direct_comparison_sqnr_nll_opt}.  

With our systematic collection of empirical data pertaining to all the above-mentioned factors, we are able to uncover patterns in the highly varied, and seemingly haphazard, effect of GPTQ on given a specific LLM quantized to a specific numerical data type. 
Here we demonstrate with the model \texttt{opt-1.3b} subject to quantization to \texttt{mxint6\_128}, \texttt{mxint4\_128}, \texttt{mxint3\_128} and \texttt{mxint2\_128} (Figure~\ref{fig:loss_landscape}, right).  
The observation is that GPTQ greatly improves \texttt{mxint3\_128} quantization, but only marginally improves its 6-bit, 4-bit and 2-bit counterparts. 
The effect of GPTQ seems highly non-monotonic as a function of quantization precision.  
Nevertheless, in the light of the underlying local loss landscape, the phenomenon can be well understood.  
First, RTN quantization to MX weight formats often lead to perturbations that are flatter than \emph{typical} radial loss profiles; the application of GPTQ, further seeks an even flatter perturbation direction in the loss landscape, as evident in Figure~\ref{fig:loss_landscape} (right).  
However, because these radial loss profiles are very \emph{step-like}, any linear or quadratic approximations typically fail to characterize them well at SNRs lower than 20 dB.  
Because of the difference in the effective radii between the RTN and GPTQ loss profiles that are both step-like, a narrow window in SNR exists within which the effect of GPTQ is substantial.  
Note that the location and size of this window is a function of the model family, the model size, and the numerical data type for weight quantization, as we described above.  

\begin{figure}[h]
    \centering
    \subfigure{
        \includegraphics[width=0.45\textwidth]{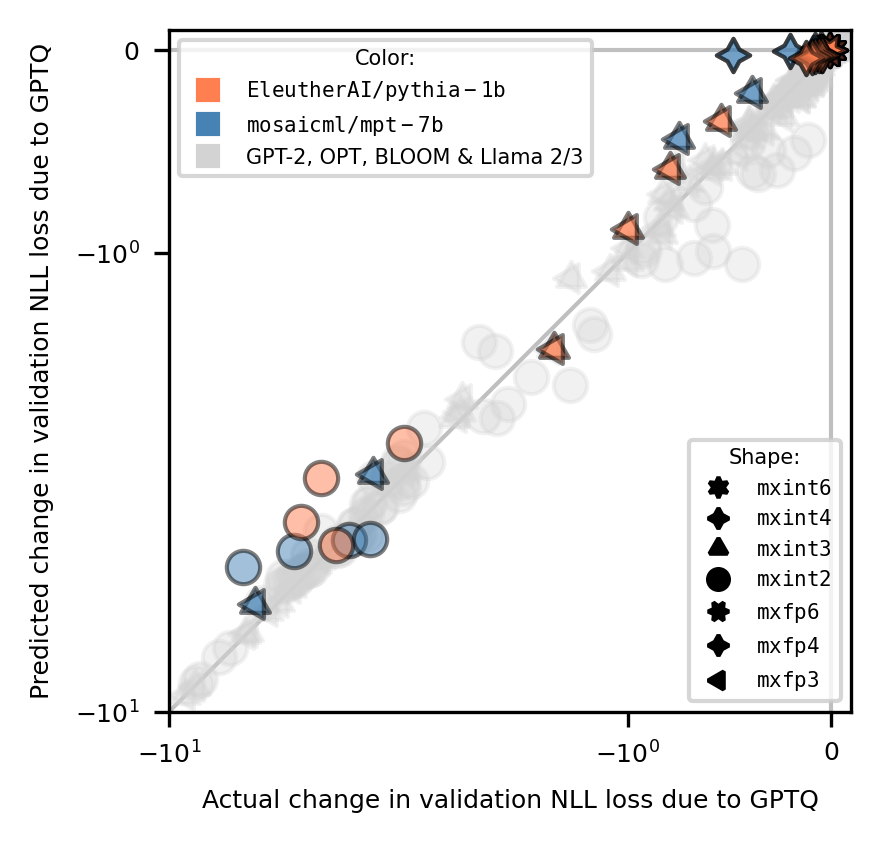}
        }
    \hfill
    \subfigure{
        \includegraphics[width=0.5\textwidth]{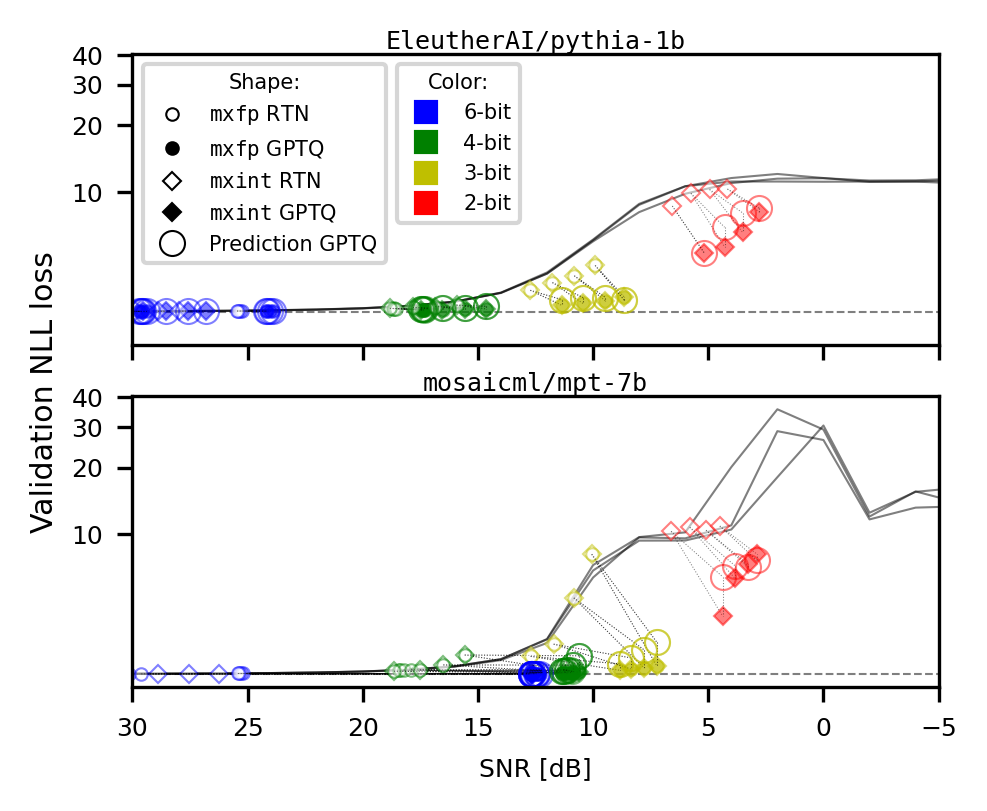}
        }
    \caption{
    Left: \textbf{A predictive model based on random forest regression.}
    Data for 18 models from the 5 LLM families used for predictive model fitting are shown in light gray; colored symbols represent held-out test data from \texttt{mpt-7b} and \texttt{pythia-1b}, respectively. 
    Prediction and observation are plotted against each other for direct comparison, diagonal line marking identity. 
    Right: \textbf{Prediction of NLL losses after GPTQ, for unseen LLMs.}
    We tested our predictive model's performance on 2 held-out LLMs from unseen model families, \texttt{mpt-7b} and \texttt{pythia-1b}.
    Convention follows Figure~\ref{fig:scaling_sqnr_nll_with_loss_landscape}, with additional large circular symbols representing model prediction of GPTQ losses. 
    }
    \label{fig:predict}
\end{figure}

\section{Building a predictive model}
\label{sec:prediction}

To sum up our findings thus far, we first found that the characteristics of local loss landscape, just like the loss itself, scales with model size in LLM families, an intrinsic model property. We also determined that choices of the low-precision data type for quantization and the PTQ process, acting within the local loss landscape, lead to different SQNRs and losses in a predictable way.

Taking these empirical rules into consideration, we now build a predictive model based on random forest regression. 
We set the hyperparameters, the number of estimators and maximum depth of the regressor, to 120 and 8, respectively.  
The regressor takes a few empirically measured features as input, and directly predicts the resulting NLL loss of the final, quantized model.  
Given a specific LLM and a specific MX data format with quantizer $Q$, the input features are: 
(a) weight parameter count $D$, (b) pre-trained loss $\nll(\vw)$, (c) SQNR of RTN quantization $\sqnr(\vw)$, (d) loss of RTN quantization $\nll\left(Q(\vw)\right)$, (e) radial slope of local loss landscape at RTN weights $\frac {d\nll} {d\sqnr} \big\vert_{Q(\vw)}$, (f) numerical format's precision \texttt{P}, number of element exponent bits \texttt{E}, and block size \texttt{K}.  
The model outputs a predicted loss after GPTQ, $\nll(Q(\vw^*))$.

We fit the model on all feature data collected from models in the 5 LLM families above, and test its prediction for 2 held-out models from unseen model families, namely \texttt{EleutherAI/pythia-1b} and \texttt{mosaicml/mpt-7b}. 
Despite the difference in model architecture, training paradigm, and even local loss landscape between \texttt{pythia-1b}, \texttt{mpt-7b}, and our existing model families, the prediction is reasonably accurate (Figure \ref{fig:predict}), suggesting that the underlying scaling laws are generalizable across both different model sizes and different LLM families. 
See Appendix~\ref{sec:prediction_interpretation} for detailed interpretation of the predictive model and salient features.

\section{Experimental procedures}
\label{sec:methods}

\subsection{Models and dataset}
We experimented with models from 5 LLM families, namely GPT-2~\citep{gpt2}, OPT~\citep{zhang2022opt}, BLOOM~\citep{workshop2023bloom}, Llama 2~\citep{touvron2023llama}, and Llama 3~\citep{llama3_2024}. 
The models were served by the Hugging Face Model Hub. 
We identify the models by their unique name string identifier throughout this paper, with their organization prefixes sometimes omitted for brevity. 

To validate the generalizability of our empirical scaling rules extracted from studying the above 5 model families, we tested their predictive power on 2 held-out LLMs, \texttt{EleutherAI/pythia-1b}~\citep{biderman2023pythia}, and \texttt{mosaicml/mpt-7b}~\citep{mpt}.  

The WikiText-2 dataset~\citep{wikitext2} was used in all experiments (see Appendix~\ref{datasets} for the generalization to other datasets), with the text tokenized by corresponding tokenizers at maximum sequence length of each respective model. 
128 examples from the training split were used as calibration dataset for PTQ algorithms. 
All examples from the validation split were used for validation.

\subsection{Numerical tensor data type and notations}
We experimented with microscaling (MX, \citealt{microscaling2023}) compliant data formats, where a block of tensor elements share a same scaling factor in the format of \texttt{e8m0} (8-bit exponent and 0-bit mantissa), and each element being of a low-precision \texttt{float} or \texttt{int} number. 
We experimented with 36 distinct MX data types with precision with block sizes ranging from 16 to 128, and element precision from 2 to 6. 

We denote MX formats by \texttt{mxfpP\_eEmM\_K} or \texttt{mxintP\_K}, following the notation from community standard~\citep{microscaling2023}, where \texttt{P} is the precision, \texttt{K} the block size, and \texttt{E}, \texttt{M} the numbers of element exponent and mantissa bits.   
For example, \texttt{mxint6\_64} represents an MX data type where the element is in \texttt{int6} and the block size 64; \texttt{mxfp4\_e2m1\_128} refers to an MX format whose element format is a custom \texttt{float4} with 1 sign bit, 2-bit exponent, 1-bit mantissa, and a block size of 128.

\subsection{GPTQ}
\label{sec:meth_ptq}
We adopted an enhanced version of GPTQ compatible with MX weight formats~\citep{sharify2024combining}, with two additional improvements. 
First, we tuned the dampening factor layerwise as a hyperparameter. 
For each layer, we did a grid search over the space $\left\{10^{-3}, 10^{-2}, \cdots, 10^3, 10^4\right\}$ and chose the dampening factor that minimized layerwise output mean squared error (MSE). 
Second, in contrast to \citet{frantar2022gptq} who performed sequential layerwise Hessian accumulation and optimization to minimize GPU memory usage, we did Hessian accumulation in unquantized network for all layers before optimization. 
In consistency with the original work, 128 sequences from the training data split was used for Hessian accumulation.

\subsection{Loss landscape mapping}
All NLL losses were evaluated on the entire validation data split at half precision.  
Second-order loss landscape features requiring backward passes, namely Hessian-vector products, were computed in single precision using the PyHessian package~\citep{yao2020pyhessian}.

\section{Conclusion}

\begin{figure*}[!hb]
  \centering
  \includegraphics[scale=0.9]{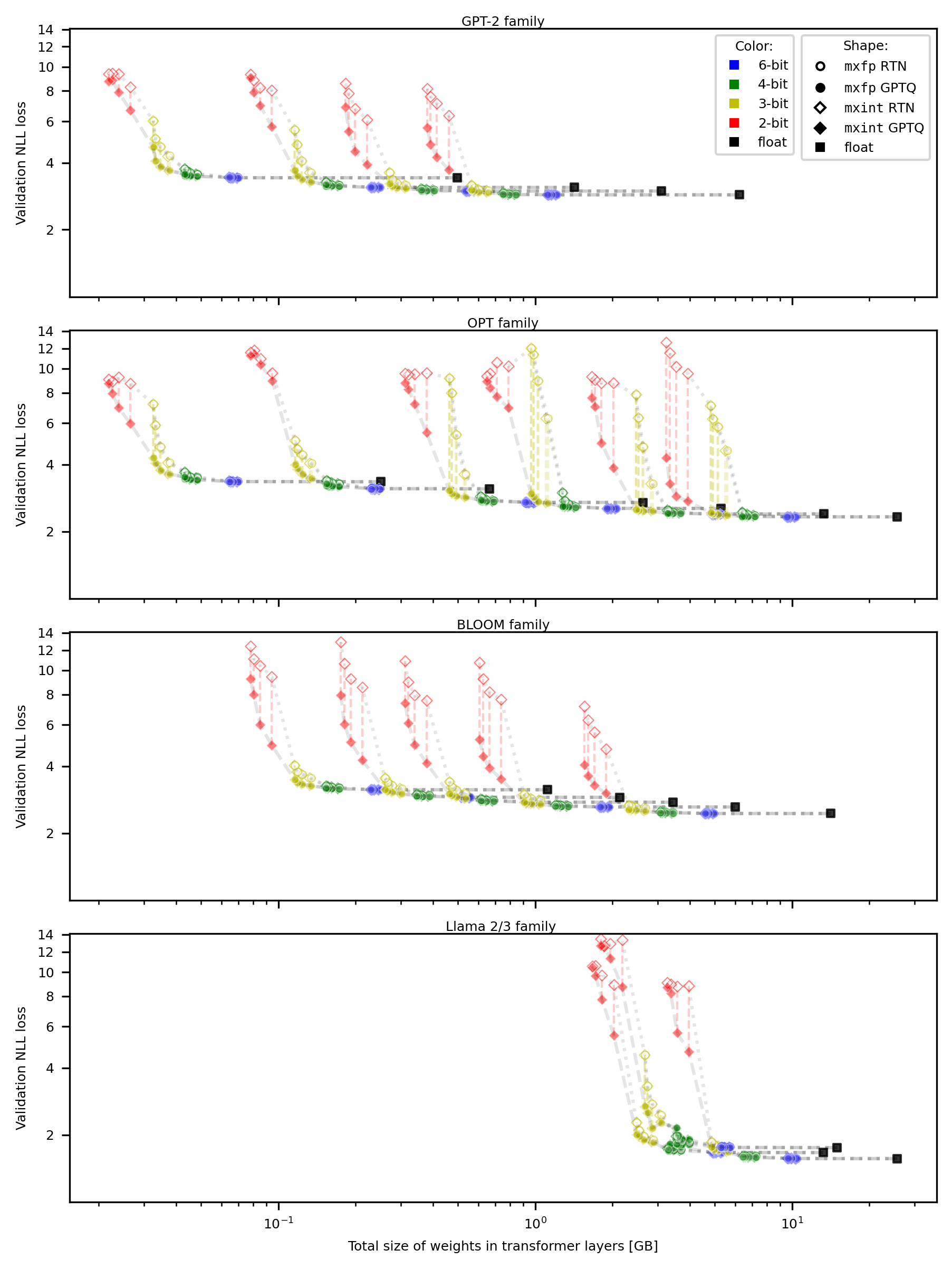} 
  \caption {
  \textbf{Tradeoff between quantized model weight size and its generalization.}
  The models in each subplot from top to bottom are: \texttt{gpt2}, \texttt{gpt2-medium}, \texttt{gpt2-large}, \texttt{gpt2-xl}; \texttt{opt-125m}, \texttt{opt-350m}, \texttt{opt-1.3b}, \texttt{opt-2.7b}, \texttt{opt-6.7b}, \texttt{opt-13b}; \texttt{bloom-560m}, \texttt{bloom-1b1}, \texttt{bloom-1b7}, \texttt{bloom-3b}, \texttt{bloom-7b1}; \texttt{Llama-2-7b-hf}, \texttt{Llama-2-13b-hf}, \texttt{Meta-Llama-3-8B}. 
  The marker colors represent different quantized precision. 
  Circles represent models quantized to \texttt{mxfp} formats, diamonds those quantized to \texttt{mxint} formats, with hollow markers standing for RTN and filled markers GPTQ. 
  Black filled squares represent the pre-trained \texttt{float} model. 
  Dashed/dotted gray lines connects the losses of the same model quantized to different data format families. 
  There are 4 such lines for each model: \texttt{mxint} (RTN): dotted, \texttt{mxfp} (RTN): dotted, \texttt{mxint} (GPTQ): dashed, and \texttt{mxfp} (GPTQ): dashed. 
  We highlight the difference before and after GPTQ by a vertical colored dashed line.
  }
  \label{fig:pareto}
\end{figure*}



In this work, we demonstrated that, just like that of pre-training, the outcome of post-training quantization of well-trained LLMs can also be predictable, thanks to underlying scaling laws governing the local loss landscape, numerical data formats and effects of PTQ algorithms. In Figure~\ref{fig:pareto}, we display the tradeoff between model quantization and quality across all models and quantization formats in this study. This graph establishes a Pareto frontier, the optimal tradeoff between larger models quantized to lower bit precisions and smaller models quantized to higher bit precisions. Moreover, since our random forest model can accurately predict NLL loss after PTQ across different model sizes and distinct model families, we argue that identifying the appropriate model size and data format for a given inference workflow is no longer a business of trial-and-error but rather one of reason guided by the underlying scaling laws of quantized LLMs. Overall, we believe our findings would provide practical value to the deployment of LLMs on resource-constrained devices.

\section{Limitations}
Due to constraint of computational resources, we experimented with models up to 13 billion parameters. 
The predictive power of our scaling rules on much larger LLMs is pending further validation.
\clearpage
\newpage
\bibliographystyle{plainnat}
\bibliography{references}

\begin{thebibliography}{44}
\providecommand{\natexlab}[1]{#1}
\providecommand{\url}[1]{\texttt{#1}}
\expandafter\ifx\csname urlstyle\endcsname\relax
  \providecommand{\doi}[1]{doi: #1}\else
  \providecommand{\doi}{doi: \begingroup \urlstyle{rm}\Url}\fi

\bibitem[Agrawal et~al.(2024)Agrawal, Hedlund, and Hechtman]{agrawal2024exmy}
Aditya Agrawal, Matthew Hedlund, and Blake Hechtman.
\newblock exmy: A data type and technique for arbitrary bit precision quantization.
\newblock 2024.

\bibitem[Alabdulmohsin et~al.(2022)Alabdulmohsin, Neyshabur, and Zhai]{alabdulmohsin2022revisiting}
Ibrahim Alabdulmohsin, Behnam Neyshabur, and Xiaohua Zhai.
\newblock Revisiting neural scaling laws in language and vision.
\newblock 2022.

\bibitem[Bahri et~al.(2024)Bahri, Dyer, Kaplan, Lee, and Sharma]{bahri2024explaining}
Yasaman Bahri, Ethan Dyer, Jared Kaplan, Jaehoon Lee, and Utkarsh Sharma.
\newblock Explaining neural scaling laws.
\newblock 2024.

\bibitem[Biderman et~al.(2023)Biderman, Schoelkopf, Anthony, Bradley, O'Brien, Hallahan, Khan, Purohit, Prashanth, Raff, Skowron, Sutawika, and van~der Wal]{biderman2023pythia}
Stella Biderman, Hailey Schoelkopf, Quentin Anthony, Herbie Bradley, Kyle O'Brien, Eric Hallahan, Mohammad~Aflah Khan, Shivanshu Purohit, USVSN~Sai Prashanth, Edward Raff, Aviya Skowron, Lintang Sutawika, and Oskar van~der Wal.
\newblock Pythia: A suite for analyzing large language models across training and scaling.
\newblock 2023.

\bibitem[Bordelon et~al.(2024)Bordelon, Atanasov, and Pehlevan]{bordelon2024dynamical}
Blake Bordelon, Alexander Atanasov, and Cengiz Pehlevan.
\newblock A dynamical model of neural scaling laws.
\newblock 2024.

\bibitem[Dettmers and Zettlemoyer(2023)]{dettmers2023case}
Tim Dettmers and Luke Zettlemoyer.
\newblock The case for 4-bit precision: k-bit inference scaling laws.
\newblock 2023.

\bibitem[Dettmers et~al.(2023)Dettmers, Pagnoni, Holtzman, and Zettlemoyer]{dettmers2023qlora}
Tim Dettmers, Artidoro Pagnoni, Ari Holtzman, and Luke Zettlemoyer.
\newblock Qlora: Efficient finetuning of quantized llms.
\newblock 2023.

\bibitem[Dodge et~al.(2021)Dodge, Sap, Marasović, Agnew, Ilharco, Groeneveld, Mitchell, and Gardner]{c4}
Jesse Dodge, Maarten Sap, Ana Marasović, William Agnew, Gabriel Ilharco, Dirk Groeneveld, Margaret Mitchell, and Matt Gardner.
\newblock Documenting large webtext corpora: A case study on the colossal clean crawled corpus, 2021.
\newblock URL \url{https://arxiv.org/abs/2104.08758}.

\bibitem[Evci et~al.(2020)Evci, Pedregosa, Gomez, and Elsen]{evci2020difficulty}
Utku Evci, Fabian Pedregosa, Aidan Gomez, and Erich Elsen.
\newblock The difficulty of training sparse neural networks.
\newblock 2020.

\bibitem[Frantar et~al.(2022)Frantar, Ashkboos, Hoefler, and Alistarh]{frantar2022gptq}
Elias Frantar, Saleh Ashkboos, Torsten Hoefler, and Dan Alistarh.
\newblock {GPTQ} accurate post-training quantization for generative pre-trained transformers.
\newblock \emph{arXiv preprint arXiv:2210.17323}, 2022.

\bibitem[Frumkin et~al.(2023)Frumkin, Gope, and Marculescu]{frumkin2023jumping}
Natalia Frumkin, Dibakar Gope, and Diana Marculescu.
\newblock Jumping through local minima: Quantization in the loss landscape of vision transformers.
\newblock 2023.

\bibitem[Gholami et~al.(2021)Gholami, Kim, Dong, Yao, Mahoney, and Keutzer]{gholami2021survey}
Amir Gholami, Sehoon Kim, Zhen Dong, Zhewei Yao, Michael~W. Mahoney, and Kurt Keutzer.
\newblock A survey of quantization methods for efficient neural network inference.
\newblock 2021.

\bibitem[Guo et~al.(2022)Guo, Zhang, Leng, Liu, Yang, Liu, Guo, and Zhu]{guo2022ant}
Cong Guo, Chen Zhang, Jingwen Leng, Zihan Liu, Fan Yang, Yunxin Liu, Minyi Guo, and Yuhao Zhu.
\newblock Ant: Exploiting adaptive numerical data type for low-bit deep neural network quantization.
\newblock 2022.

\bibitem[Henighan et~al.(2020)Henighan, Kaplan, Katz, Chen, Hesse, Jackson, Jun, Brown, Dhariwal, Gray, Hallacy, Mann, Radford, Ramesh, Ryder, Ziegler, Schulman, Amodei, and McCandlish]{henighan2020scaling}
Tom Henighan, Jared Kaplan, Mor Katz, Mark Chen, Christopher Hesse, Jacob Jackson, Heewoo Jun, Tom~B. Brown, Prafulla Dhariwal, Scott Gray, Chris Hallacy, Benjamin Mann, Alec Radford, Aditya Ramesh, Nick Ryder, Daniel~M. Ziegler, John Schulman, Dario Amodei, and Sam McCandlish.
\newblock Scaling laws for autoregressive generative modeling.
\newblock 2020.

\bibitem[Hu et~al.(2022)Hu, Meinel, and Yang]{hu2022empirical}
Ting Hu, Christoph Meinel, and Haojin Yang.
\newblock Empirical evaluation of post-training quantization methods for language tasks.
\newblock 2022.

\bibitem[Huang et~al.(2024)Huang, Ma, Qin, Zheng, Lv, Chen, Luo, Qi, Liu, and Magno]{huang2024good}
Wei Huang, Xudong Ma, Haotong Qin, Xingyu Zheng, Chengtao Lv, Hong Chen, Jie Luo, Xiaojuan Qi, Xianglong Liu, and Michele Magno.
\newblock How good are low-bit quantized llama3 models? an empirical study.
\newblock 2024.

\bibitem[Jeon et~al.(2024)Jeon, Kim, and joon Kim]{jeon2024l4q}
Hyesung Jeon, Yulhwa Kim, and Jae joon Kim.
\newblock L4q: Parameter efficient quantization-aware fine-tuning on large language models, 2024.

\bibitem[Kaplan et~al.(2020)Kaplan, McCandlish, Henighan, Brown, Chess, Child, Gray, Radford, Wu, and Amodei]{kaplan2020scaling}
Jared Kaplan, Sam McCandlish, Tom Henighan, Tom~B. Brown, Benjamin Chess, Rewon Child, Scott Gray, Alec Radford, Jeffrey Wu, and Dario Amodei.
\newblock Scaling laws for neural language models.
\newblock 2020.

\bibitem[Kim et~al.(2024)Kim, Hooper, Gholami, Dong, Li, Shen, Mahoney, and Keutzer]{kim2024squeezellm}
Sehoon Kim, Coleman Hooper, Amir Gholami, Zhen Dong, Xiuyu Li, Sheng Shen, Michael~W. Mahoney, and Kurt Keutzer.
\newblock Squeezellm: Dense-and-sparse quantization.
\newblock 2024.

\bibitem[Kim et~al.(2023)Kim, Henry, Fahim, and Awadalla]{kim2023finequant}
Young~Jin Kim, Rawn Henry, Raffy Fahim, and Hany~Hassan Awadalla.
\newblock Finequant: Unlocking efficiency with fine-grained weight-only quantization for llms.
\newblock 2023.

\bibitem[Lee et~al.(2024)Lee, Jin, Kim, Kim, and Park]{lee2024owq}
Changhun Lee, Jungyu Jin, Taesu Kim, Hyungjun Kim, and Eunhyeok Park.
\newblock Owq: Outlier-aware weight quantization for efficient fine-tuning and inference of large language models.
\newblock 2024.

\bibitem[Li et~al.(2023)Li, Yu, Liang, He, Karampatziakis, Chen, and Zhao]{li2023loftq}
Yixiao Li, Yifan Yu, Chen Liang, Pengcheng He, Nikos Karampatziakis, Weizhu Chen, and Tuo Zhao.
\newblock Loftq: Lora-fine-tuning-aware quantization for large language models, 2023.

\bibitem[Lin et~al.(2024)Lin, Tang, Tang, Yang, Chen, Wang, Xiao, Dang, Gan, and Han]{lin2024awq}
Ji~Lin, Jiaming Tang, Haotian Tang, Shang Yang, Wei-Ming Chen, Wei-Chen Wang, Guangxuan Xiao, Xingyu Dang, Chuang Gan, and Song Han.
\newblock Awq: Activation-aware weight quantization for llm compression and acceleration.
\newblock 2024.

\bibitem[Louppe et~al.(2013)Louppe, Wehenkel, Sutera, and Geurts]{gini}
Gilles Louppe, Louis Wehenkel, Antonio Sutera, and Pierre Geurts.
\newblock Understanding variable importances in forests of randomized trees.
\newblock volume~26, 12 2013.

\bibitem[Merity et~al.(2016)Merity, Xiong, Bradbury, and Socher]{wikitext2}
Stephen Merity, Caiming Xiong, James Bradbury, and Richard Socher.
\newblock Pointer sentinel mixture models.
\newblock \emph{arXiv preprint arXiv:1609.07843}, 2016.

\bibitem[Meta(2024)]{llama3_2024}
Meta.
\newblock Introducing meta llama 3: the most capable openly available llm to date 2024, 2024.
\newblock URL \url{https://ai.meta.com/blog/meta-llama-3/}.

\bibitem[MosaicML(2023)]{mpt}
MosaicML.
\newblock Introducing mpt-7b: A new standard for open-source, commercially usable llms, 2023.
\newblock URL \url{https://www.databricks.com/blog/mpt-7b}.

\bibitem[Muennighoff et~al.(2023)Muennighoff, Rush, Barak, Scao, Piktus, Tazi, Pyysalo, Wolf, and Raffel]{muennighoff2023scaling}
Niklas Muennighoff, Alexander~M. Rush, Boaz Barak, Teven~Le Scao, Aleksandra Piktus, Nouamane Tazi, Sampo Pyysalo, Thomas Wolf, and Colin Raffel.
\newblock Scaling data-constrained language models.
\newblock 2023.

\bibitem[Nahshan et~al.(2020)Nahshan, Chmiel, Baskin, Zheltonozhskii, Banner, Bronstein, and Mendelson]{nahshan2020loss}
Yury Nahshan, Brian Chmiel, Chaim Baskin, Evgenii Zheltonozhskii, Ron Banner, Alex~M. Bronstein, and Avi Mendelson.
\newblock Loss aware post-training quantization.
\newblock 2020.

\bibitem[Paperno et~al.(2016)Paperno, Kruszewski, Lazaridou, Pham, Bernardi, Pezzelle, Baroni, Boleda, and Fernández]{lambada}
Denis Paperno, Germán Kruszewski, Angeliki Lazaridou, Quan~Ngoc Pham, Raffaella Bernardi, Sandro Pezzelle, Marco Baroni, Gemma Boleda, and Raquel Fernández.
\newblock The lambada dataset: Word prediction requiring a broad discourse context, 2016.
\newblock URL \url{https://arxiv.org/abs/1606.06031}.

\bibitem[Park et~al.(2024)Park, Park, Kim, Lee, Kim, Kwon, Kwon, Kim, Lee, and Lee]{park2024lutgemm}
Gunho Park, Baeseong Park, Minsub Kim, Sungjae Lee, Jeonghoon Kim, Beomseok Kwon, Se~Jung Kwon, Byeongwook Kim, Youngjoo Lee, and Dongsoo Lee.
\newblock Lut-gemm: Quantized matrix multiplication based on luts for efficient inference in large-scale generative language models.
\newblock 2024.

\bibitem[Radford et~al.(2019)Radford, Wu, Child, Luan, Amodei, Sutskever, et~al.]{gpt2}
Alec Radford, Jeffrey Wu, Rewon Child, David Luan, Dario Amodei, Ilya Sutskever, et~al.
\newblock Language models are unsupervised multitask learners.
\newblock \emph{OpenAI blog}, 1\penalty0 (8):\penalty0 9, 2019.

\bibitem[Rouhani et~al.(2023)Rouhani, Garegrat, Savell, More, Han, Zhao, Hall, Klar, Chung, Yu, Schulte, Wittig, Bratt, Stephens, Milanovic, Brothers, Dubey, Cornea, Heinecke, Rodriguez, Langhammer, Deng, Naumov, Micikevicius, Siu, and Verrilli]{microscaling2023}
Bita~Darvish Rouhani, Nitin Garegrat, Tom Savell, Ankit More, Kyung-Nam Han, Ritchie Zhao, Mathew Hall, Jasmine Klar, Eric Chung, Yuan Yu, Michael Schulte, Ralph Wittig, Ian Bratt, Nigel Stephens, Jelena Milanovic, John Brothers, Pradeep Dubey, Marius Cornea, Alexander Heinecke, Andres Rodriguez, Martin Langhammer, Summer Deng, Maxim Naumov, Paulius Micikevicius, Michael Siu, and Colin Verrilli.
\newblock Ocp microscaling formats (mx) specification.
\newblock \emph{Open Compute Project}, 2023.

\bibitem[Sharify et~al.(2024)Sharify, Xu, Yazar, and Wang]{sharify2024combining}
Sayeh Sharify, Zifei Xu, Wanzin Yazar, and Xin Wang.
\newblock Combining multiple post-training techniques to achieve most efficient quantized llms.
\newblock 2024.

\bibitem[Song et~al.(2024)Song, Liu, Tegmark, and Gore]{song2024resource}
Jinyeop Song, Ziming Liu, Max Tegmark, and Jeff Gore.
\newblock A resource model for neural scaling law.
\newblock 2024.

\bibitem[Su et~al.(2024)Su, Tian, Shen, and Cai]{su2024unraveling}
Hui Su, Zhi Tian, Xiaoyu Shen, and Xunliang Cai.
\newblock Unraveling the mystery of scaling laws: Part i.
\newblock 2024.

\bibitem[Touvron et~al.(2023)Touvron, Martin, Stone, Albert, Almahairi, Babaei, Bashlykov, Batra, Bhargava, Bhosale, Bikel, Blecher, Ferrer, Chen, Cucurull, Esiobu, Fernandes, Fu, Fu, Fuller, Gao, Goswami, Goyal, Hartshorn, Hosseini, Hou, Inan, Kardas, Kerkez, Khabsa, Kloumann, Korenev, Koura, Lachaux, Lavril, Lee, Liskovich, Lu, Mao, Martinet, Mihaylov, Mishra, Molybog, Nie, Poulton, Reizenstein, Rungta, Saladi, Schelten, Silva, Smith, Subramanian, Tan, Tang, Taylor, Williams, Kuan, Xu, Yan, Zarov, Zhang, Fan, Kambadur, Narang, Rodriguez, Stojnic, Edunov, and Scialom]{touvron2023llama}
Hugo Touvron, Louis Martin, Kevin Stone, Peter Albert, Amjad Almahairi, Yasmine Babaei, Nikolay Bashlykov, Soumya Batra, Prajjwal Bhargava, Shruti Bhosale, Dan Bikel, Lukas Blecher, Cristian~Canton Ferrer, Moya Chen, Guillem Cucurull, David Esiobu, Jude Fernandes, Jeremy Fu, Wenyin Fu, Brian Fuller, Cynthia Gao, Vedanuj Goswami, Naman Goyal, Anthony Hartshorn, Saghar Hosseini, Rui Hou, Hakan Inan, Marcin Kardas, Viktor Kerkez, Madian Khabsa, Isabel Kloumann, Artem Korenev, Punit~Singh Koura, Marie-Anne Lachaux, Thibaut Lavril, Jenya Lee, Diana Liskovich, Yinghai Lu, Yuning Mao, Xavier Martinet, Todor Mihaylov, Pushkar Mishra, Igor Molybog, Yixin Nie, Andrew Poulton, Jeremy Reizenstein, Rashi Rungta, Kalyan Saladi, Alan Schelten, Ruan Silva, Eric~Michael Smith, Ranjan Subramanian, Xiaoqing~Ellen Tan, Binh Tang, Ross Taylor, Adina Williams, Jian~Xiang Kuan, Puxin Xu, Zheng Yan, Iliyan Zarov, Yuchen Zhang, Angela Fan, Melanie Kambadur, Sharan Narang, Aurelien Rodriguez, Robert Stojnic, Sergey Edunov, and Thomas
  Scialom.
\newblock Llama 2: Open foundation and fine-tuned chat models.
\newblock \emph{arXiv preprint arXiv:2307.09288}, 2023.

\bibitem[Vaswani et~al.(2023)Vaswani, Shazeer, Parmar, Uszkoreit, Jones, Gomez, Kaiser, and Polosukhin]{vaswani2023attention}
Ashish Vaswani, Noam Shazeer, Niki Parmar, Jakob Uszkoreit, Llion Jones, Aidan~N. Gomez, Lukasz Kaiser, and Illia Polosukhin.
\newblock Attention is all you need, 2023.

\bibitem[Workshop et~al.(2023)Workshop, :, Scao, Fan, Akiki, Pavlick, Ilić, Hesslow, Castagné, Luccioni, Yvon, Gallé, Tow, Rush, Biderman, Webson, Ammanamanchi, Wang, Sagot, Muennighoff, del Moral, Ruwase, Bawden, Bekman, McMillan-Major, Beltagy, Nguyen, Saulnier, Tan, Suarez, Sanh, Laurençon, Jernite, Launay, Mitchell, Raffel, Gokaslan, Simhi, Soroa, Aji, Alfassy, Rogers, Nitzav, Xu, Mou, Emezue, Klamm, Leong, van Strien, Adelani, Radev, Ponferrada, Levkovizh, Kim, Natan, Toni, Dupont, Kruszewski, Pistilli, Elsahar, Benyamina, Tran, Yu, Abdulmumin, Johnson, Gonzalez-Dios, de~la Rosa, Chim, Dodge, Zhu, Chang, Frohberg, Tobing, Bhattacharjee, Almubarak, Chen, Lo, Werra, Weber, Phan, allal, Tanguy, Dey, Muñoz, Masoud, Grandury, Šaško, Huang, Coavoux, Singh, Jiang, Vu, Jauhar, Ghaleb, Subramani, Kassner, Khamis, Nguyen, Espejel, de~Gibert, Villegas, Henderson, Colombo, Amuok, Lhoest, Harliman, Bommasani, López, Ribeiro, Osei, Pyysalo, Nagel, Bose, Muhammad, Sharma, Longpre, Nikpoor, Silberberg, Pai,
  Zink, Torrent, Schick, Thrush, Danchev, Nikoulina, Laippala, Lepercq, Prabhu, Alyafeai, Talat, Raja, Heinzerling, Si, Taşar, Salesky, Mielke, Lee, Sharma, Santilli, Chaffin, Stiegler, Datta, Szczechla, Chhablani, Wang, Pandey, Strobelt, Fries, Rozen, Gao, Sutawika, Bari, Al-shaibani, Manica, Nayak, Teehan, Albanie, Shen, Ben-David, Bach, Kim, Bers, Fevry, Neeraj, Thakker, Raunak, Tang, Yong, Sun, Brody, Uri, Tojarieh, Roberts, Chung, Tae, Phang, Press, Li, Narayanan, Bourfoune, Casper, Rasley, Ryabinin, Mishra, Zhang, Shoeybi, Peyrounette, Patry, Tazi, Sanseviero, von Platen, Cornette, Lavallée, Lacroix, Rajbhandari, Gandhi, Smith, Requena, Patil, Dettmers, Baruwa, Singh, Cheveleva, Ligozat, Subramonian, Névéol, Lovering, Garrette, Tunuguntla, Reiter, Taktasheva, Voloshina, Bogdanov, Winata, Schoelkopf, Kalo, Novikova, Forde, Clive, Kasai, Kawamura, Hazan, Carpuat, Clinciu, Kim, Cheng, Serikov, Antverg, van~der Wal, Zhang, Zhang, Gehrmann, Mirkin, Pais, Shavrina, Scialom, Yun, Limisiewicz, Rieser,
  Protasov, Mikhailov, Pruksachatkun, Belinkov, Bamberger, Kasner, Rueda, Pestana, Feizpour, Khan, Faranak, Santos, Hevia, Unldreaj, Aghagol, Abdollahi, Tammour, HajiHosseini, Behroozi, Ajibade, Saxena, Ferrandis, McDuff, Contractor, Lansky, David, Kiela, Nguyen, Tan, Baylor, Ozoani, Mirza, Ononiwu, Rezanejad, Jones, Bhattacharya, Solaiman, Sedenko, Nejadgholi, Passmore, Seltzer, Sanz, Dutra, Samagaio, Elbadri, Mieskes, Gerchick, Akinlolu, McKenna, Qiu, Ghauri, Burynok, Abrar, Rajani, Elkott, Fahmy, Samuel, An, Kromann, Hao, Alizadeh, Shubber, Wang, Roy, Viguier, Le, Oyebade, Le, Yang, Nguyen, Kashyap, Palasciano, Callahan, Shukla, Miranda-Escalada, Singh, Beilharz, Wang, Brito, Zhou, Jain, Xu, Fourrier, Periñán, Molano, Yu, Manjavacas, Barth, Fuhrimann, Altay, Bayrak, Burns, Vrabec, Bello, Dash, Kang, Giorgi, Golde, Posada, Sivaraman, Bulchandani, Liu, Shinzato, de~Bykhovetz, Takeuchi, Pàmies, Castillo, Nezhurina, Sänger, Samwald, Cullan, Weinberg, Wolf, Mihaljcic, Liu, Freidank, Kang, Seelam, Dahlberg,
  Broad, Muellner, Fung, Haller, Chandrasekhar, Eisenberg, Martin, Canalli, Su, Su, Cahyawijaya, Garda, Deshmukh, Mishra, Kiblawi, Ott, Sang-aroonsiri, Kumar, Schweter, Bharati, Laud, Gigant, Kainuma, Kusa, Labrak, Bajaj, Venkatraman, Xu, Xu, Xu, Tan, Xie, Ye, Bras, Belkada, and Wolf]{workshop2023bloom}
BigScience Workshop, :, Teven~Le Scao, Angela Fan, Christopher Akiki, Ellie Pavlick, Suzana Ilić, Daniel Hesslow, Roman Castagné, Alexandra~Sasha Luccioni, François Yvon, Matthias Gallé, Jonathan Tow, Alexander~M. Rush, Stella Biderman, Albert Webson, Pawan~Sasanka Ammanamanchi, Thomas Wang, Benoît Sagot, Niklas Muennighoff, Albert~Villanova del Moral, Olatunji Ruwase, Rachel Bawden, Stas Bekman, Angelina McMillan-Major, Iz~Beltagy, Huu Nguyen, Lucile Saulnier, Samson Tan, Pedro~Ortiz Suarez, Victor Sanh, Hugo Laurençon, Yacine Jernite, Julien Launay, Margaret Mitchell, Colin Raffel, Aaron Gokaslan, Adi Simhi, Aitor Soroa, Alham~Fikri Aji, Amit Alfassy, Anna Rogers, Ariel~Kreisberg Nitzav, Canwen Xu, Chenghao Mou, Chris Emezue, Christopher Klamm, Colin Leong, Daniel van Strien, David~Ifeoluwa Adelani, Dragomir Radev, Eduardo~González Ponferrada, Efrat Levkovizh, Ethan Kim, Eyal~Bar Natan, Francesco~De Toni, Gérard Dupont, Germán Kruszewski, Giada Pistilli, Hady Elsahar, Hamza Benyamina, Hieu Tran,
  Ian Yu, Idris Abdulmumin, Isaac Johnson, Itziar Gonzalez-Dios, Javier de~la Rosa, Jenny Chim, Jesse Dodge, Jian Zhu, Jonathan Chang, Jörg Frohberg, Joseph Tobing, Joydeep Bhattacharjee, Khalid Almubarak, Kimbo Chen, Kyle Lo, Leandro~Von Werra, Leon Weber, Long Phan, Loubna~Ben allal, Ludovic Tanguy, Manan Dey, Manuel~Romero Muñoz, Maraim Masoud, María Grandury, Mario Šaško, Max Huang, Maximin Coavoux, Mayank Singh, Mike Tian-Jian Jiang, Minh~Chien Vu, Mohammad~A. Jauhar, Mustafa Ghaleb, Nishant Subramani, Nora Kassner, Nurulaqilla Khamis, Olivier Nguyen, Omar Espejel, Ona de~Gibert, Paulo Villegas, Peter Henderson, Pierre Colombo, Priscilla Amuok, Quentin Lhoest, Rheza Harliman, Rishi Bommasani, Roberto~Luis López, Rui Ribeiro, Salomey Osei, Sampo Pyysalo, Sebastian Nagel, Shamik Bose, Shamsuddeen~Hassan Muhammad, Shanya Sharma, Shayne Longpre, Somaieh Nikpoor, Stanislav Silberberg, Suhas Pai, Sydney Zink, Tiago~Timponi Torrent, Timo Schick, Tristan Thrush, Valentin Danchev, Vassilina Nikoulina,
  Veronika Laippala, Violette Lepercq, Vrinda Prabhu, Zaid Alyafeai, Zeerak Talat, Arun Raja, Benjamin Heinzerling, Chenglei Si, Davut~Emre Taşar, Elizabeth Salesky, Sabrina~J. Mielke, Wilson~Y. Lee, Abheesht Sharma, Andrea Santilli, Antoine Chaffin, Arnaud Stiegler, Debajyoti Datta, Eliza Szczechla, Gunjan Chhablani, Han Wang, Harshit Pandey, Hendrik Strobelt, Jason~Alan Fries, Jos Rozen, Leo Gao, Lintang Sutawika, M~Saiful Bari, Maged~S. Al-shaibani, Matteo Manica, Nihal Nayak, Ryan Teehan, Samuel Albanie, Sheng Shen, Srulik Ben-David, Stephen~H. Bach, Taewoon Kim, Tali Bers, Thibault Fevry, Trishala Neeraj, Urmish Thakker, Vikas Raunak, Xiangru Tang, Zheng-Xin Yong, Zhiqing Sun, Shaked Brody, Yallow Uri, Hadar Tojarieh, Adam Roberts, Hyung~Won Chung, Jaesung Tae, Jason Phang, Ofir Press, Conglong Li, Deepak Narayanan, Hatim Bourfoune, Jared Casper, Jeff Rasley, Max Ryabinin, Mayank Mishra, Minjia Zhang, Mohammad Shoeybi, Myriam Peyrounette, Nicolas Patry, Nouamane Tazi, Omar Sanseviero, Patrick von
  Platen, Pierre Cornette, Pierre~François Lavallée, Rémi Lacroix, Samyam Rajbhandari, Sanchit Gandhi, Shaden Smith, Stéphane Requena, Suraj Patil, Tim Dettmers, Ahmed Baruwa, Amanpreet Singh, Anastasia Cheveleva, Anne-Laure Ligozat, Arjun Subramonian, Aurélie Névéol, Charles Lovering, Dan Garrette, Deepak Tunuguntla, Ehud Reiter, Ekaterina Taktasheva, Ekaterina Voloshina, Eli Bogdanov, Genta~Indra Winata, Hailey Schoelkopf, Jan-Christoph Kalo, Jekaterina Novikova, Jessica~Zosa Forde, Jordan Clive, Jungo Kasai, Ken Kawamura, Liam Hazan, Marine Carpuat, Miruna Clinciu, Najoung Kim, Newton Cheng, Oleg Serikov, Omer Antverg, Oskar van~der Wal, Rui Zhang, Ruochen Zhang, Sebastian Gehrmann, Shachar Mirkin, Shani Pais, Tatiana Shavrina, Thomas Scialom, Tian Yun, Tomasz Limisiewicz, Verena Rieser, Vitaly Protasov, Vladislav Mikhailov, Yada Pruksachatkun, Yonatan Belinkov, Zachary Bamberger, Zdeněk Kasner, Alice Rueda, Amanda Pestana, Amir Feizpour, Ammar Khan, Amy Faranak, Ana Santos, Anthony Hevia, Antigona
  Unldreaj, Arash Aghagol, Arezoo Abdollahi, Aycha Tammour, Azadeh HajiHosseini, Bahareh Behroozi, Benjamin Ajibade, Bharat Saxena, Carlos~Muñoz Ferrandis, Daniel McDuff, Danish Contractor, David Lansky, Davis David, Douwe Kiela, Duong~A. Nguyen, Edward Tan, Emi Baylor, Ezinwanne Ozoani, Fatima Mirza, Frankline Ononiwu, Habib Rezanejad, Hessie Jones, Indrani Bhattacharya, Irene Solaiman, Irina Sedenko, Isar Nejadgholi, Jesse Passmore, Josh Seltzer, Julio~Bonis Sanz, Livia Dutra, Mairon Samagaio, Maraim Elbadri, Margot Mieskes, Marissa Gerchick, Martha Akinlolu, Michael McKenna, Mike Qiu, Muhammed Ghauri, Mykola Burynok, Nafis Abrar, Nazneen Rajani, Nour Elkott, Nour Fahmy, Olanrewaju Samuel, Ran An, Rasmus Kromann, Ryan Hao, Samira Alizadeh, Sarmad Shubber, Silas Wang, Sourav Roy, Sylvain Viguier, Thanh Le, Tobi Oyebade, Trieu Le, Yoyo Yang, Zach Nguyen, Abhinav~Ramesh Kashyap, Alfredo Palasciano, Alison Callahan, Anima Shukla, Antonio Miranda-Escalada, Ayush Singh, Benjamin Beilharz, Bo~Wang, Caio Brito,
  Chenxi Zhou, Chirag Jain, Chuxin Xu, Clémentine Fourrier, Daniel~León Periñán, Daniel Molano, Dian Yu, Enrique Manjavacas, Fabio Barth, Florian Fuhrimann, Gabriel Altay, Giyaseddin Bayrak, Gully Burns, Helena~U. Vrabec, Imane Bello, Ishani Dash, Jihyun Kang, John Giorgi, Jonas Golde, Jose~David Posada, Karthik~Rangasai Sivaraman, Lokesh Bulchandani, Lu~Liu, Luisa Shinzato, Madeleine~Hahn de~Bykhovetz, Maiko Takeuchi, Marc Pàmies, Maria~A Castillo, Marianna Nezhurina, Mario Sänger, Matthias Samwald, Michael Cullan, Michael Weinberg, Michiel~De Wolf, Mina Mihaljcic, Minna Liu, Moritz Freidank, Myungsun Kang, Natasha Seelam, Nathan Dahlberg, Nicholas~Michio Broad, Nikolaus Muellner, Pascale Fung, Patrick Haller, Ramya Chandrasekhar, Renata Eisenberg, Robert Martin, Rodrigo Canalli, Rosaline Su, Ruisi Su, Samuel Cahyawijaya, Samuele Garda, Shlok~S Deshmukh, Shubhanshu Mishra, Sid Kiblawi, Simon Ott, Sinee Sang-aroonsiri, Srishti Kumar, Stefan Schweter, Sushil Bharati, Tanmay Laud, Théo Gigant, Tomoya
  Kainuma, Wojciech Kusa, Yanis Labrak, Yash~Shailesh Bajaj, Yash Venkatraman, Yifan Xu, Yingxin Xu, Yu~Xu, Zhe Tan, Zhongli Xie, Zifan Ye, Mathilde Bras, Younes Belkada, and Thomas Wolf.
\newblock Bloom: A 176b-parameter open-access multilingual language model, 2023.

\bibitem[Xiao et~al.(2024)Xiao, Lin, Seznec, Wu, Demouth, and Han]{xiao2024smoothquant}
Guangxuan Xiao, Ji~Lin, Mickael Seznec, Hao Wu, Julien Demouth, and Song Han.
\newblock Smoothquant: Accurate and efficient post-training quantization for large language models.
\newblock 2024.

\bibitem[Yao et~al.(2020)Yao, Gholami, Keutzer, and Mahoney]{yao2020pyhessian}
Zhewei Yao, Amir Gholami, Kurt Keutzer, and Michael Mahoney.
\newblock Pyhessian: Neural networks through the lens of the hessian, 2020.

\bibitem[Yao et~al.(2022)Yao, Aminabadi, Zhang, Wu, Li, and He]{yao2022zeroquant}
Zhewei Yao, Reza~Yazdani Aminabadi, Minjia Zhang, Xiaoxia Wu, Conglong Li, and Yuxiong He.
\newblock Zeroquant: Efficient and affordable post-training quantization for large-scale transformers.
\newblock 2022.

\bibitem[Yuan et~al.(2023)Yuan, Liu, Wu, Yang, Wu, Sun, Liu, Wang, and Wu]{yuan2023benchmarking}
Zhihang Yuan, Jiawei Liu, Jiaxiang Wu, Dawei Yang, Qiang Wu, Guangyu Sun, Wenyu Liu, Xinggang Wang, and Bingzhe Wu.
\newblock Benchmarking the reliability of post-training quantization: a particular focus on worst-case performance.
\newblock 2023.

\bibitem[Zhang et~al.(2022)Zhang, Roller, Goyal, Artetxe, Chen, Chen, Dewan, Diab, Li, Lin, Mihaylov, Ott, Shleifer, Shuster, Simig, Koura, Sridhar, Wang, and Zettlemoyer]{zhang2022opt}
Susan Zhang, Stephen Roller, Naman Goyal, Mikel Artetxe, Moya Chen, Shuohui Chen, Christopher Dewan, Mona Diab, Xian Li, Xi~Victoria Lin, Todor Mihaylov, Myle Ott, Sam Shleifer, Kurt Shuster, Daniel Simig, Punit~Singh Koura, Anjali Sridhar, Tianlu Wang, and Luke Zettlemoyer.
\newblock {OPT}: Open pre-trained transformer language models.
\newblock \emph{arXiv preprint arXiv:2205.01068}, 2022.

\end{thebibliography}
\newpage
\clearpage
\appendix

\section{Scaling of $\ell_2$-norms of model weights}
\label{sec:weight_norm}

\begin{figure}[h!]
  \centering
  \includegraphics[scale=1.0]{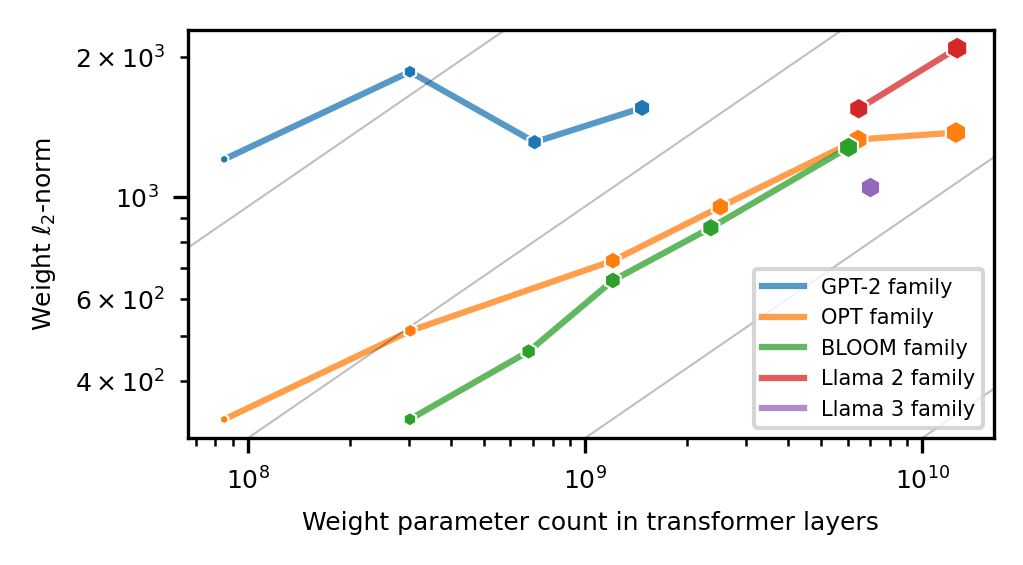}
  \caption{
    \textbf{Scaling of weight $\ell_2$-norm.}
    Convention same as in Figure~\ref{fig:loss_scaling_and_landscape} Left.
    Light gray lines in the background mark square-root power laws, $\norm{\vw} \propto D^{\frac 1 2}$.  
  }
  \label{fig:weight_norm_scaling}
\end{figure}

In Figure~\ref{fig:weight_norm_scaling}, we summarize the scaling of the $\ell_2$-norms of transformer weights, for all models in the 5 LLM families under study.
We found that, with the exception of the GPT-2 and OPT families, $\norm{\vw}$ scales close to half power laws w.r.t. parameter count $D$, suggesting a rather constant element-wise weight magnitude across models of different sizes.  
We also found that, not surprisingly, the closeness to half power law scaling of $\ell_2$-norms is correlated with the constancy of SQNRs for all MX data types across models.  

\section{SQNR and NLL of MX formats}
\label{sec:data_format}
\begin{figure*}[h!]
  \centering
  \includegraphics[scale=0.8]{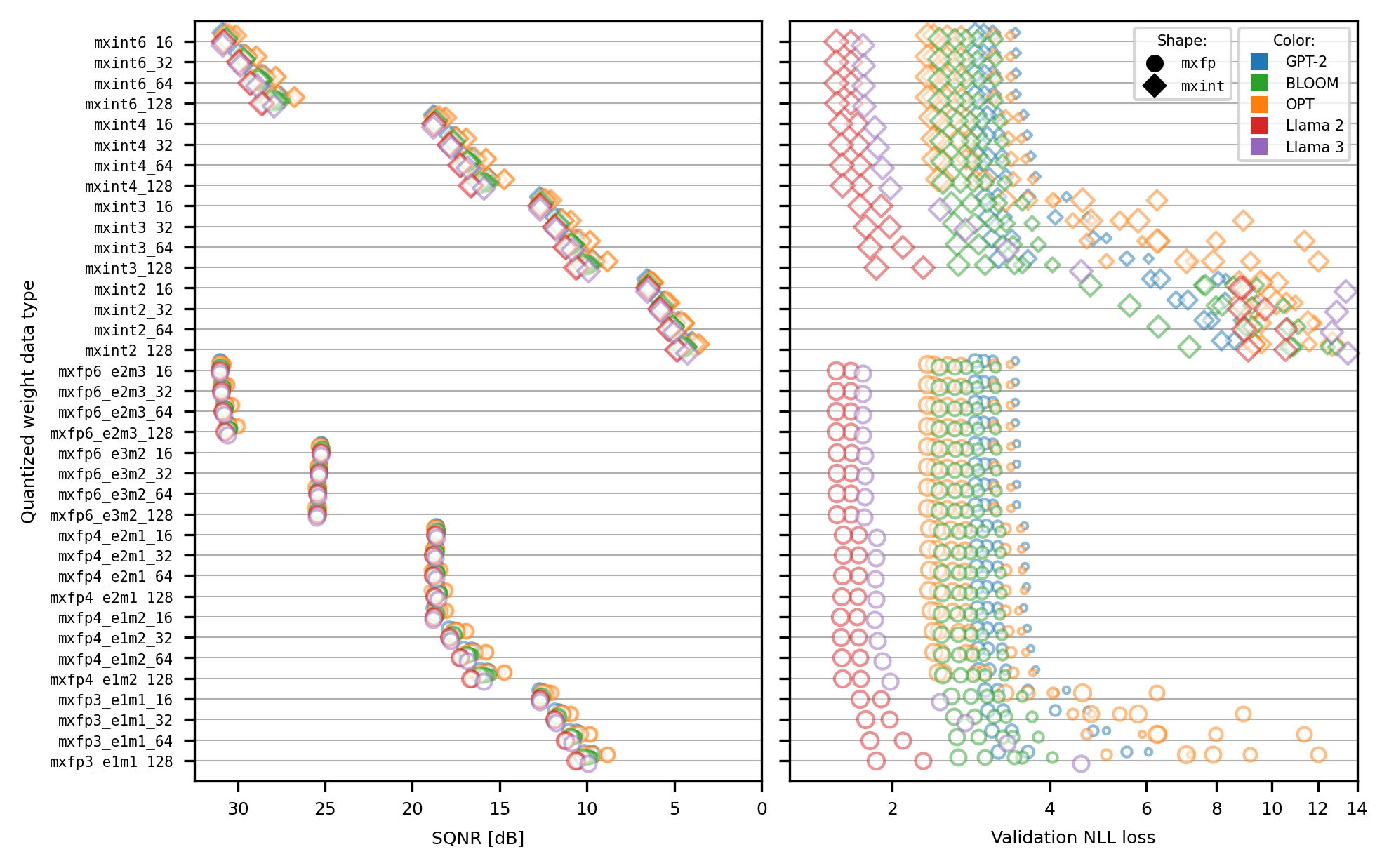} 
  \caption {
    \textbf{SQNRs and NLL losses resulting from weight quantization, before PTQ.}  
    We show round-to-nearest (RTN) results for all models in multiple LLM families. 
    Consistent with convention set in Figure~\ref{fig:loss_scaling_and_landscape} (left), model families are color-coded and model sizes are encoded by symbol sizes. 
  }
  \label{fig:rtn_sqnr_nll}
\end{figure*}

\begin{figure*}[h!]
  \centering
  \includegraphics[scale=0.8]{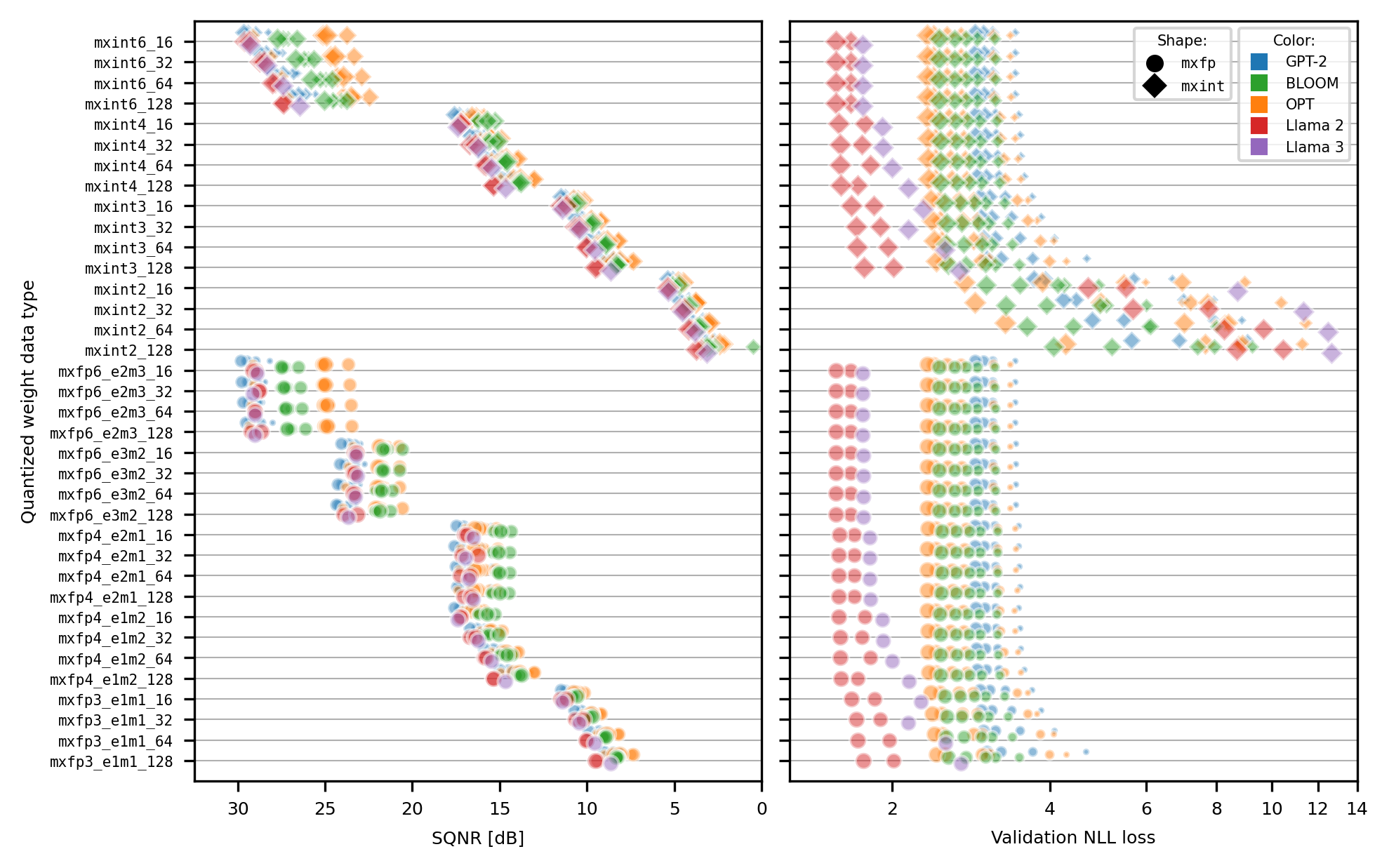} 
  \caption {
    \textbf{SQNRs and NLL losses resulting from weight quantization, after PTQ.}  
    Similar to Figure~\ref{fig:rtn_sqnr_nll}, we show GPTQ results for all models in multiple LLM families. 
  }
  \label{fig:gptq_sqnr_nll}
\end{figure*}
\section{Scaling in the case of PTQ to traditional \texttt{int} quantization}
\label{sec:int_quantization}

We note that, in the case of traditional weight quantization to integer (\texttt{int}) numerical formats, an extra step of calibration is necessary.  
Calibration optimizes additional parameters per quantizer, namely a scale and/or a zero point, depending on the quantization scheme. 
The affine transformation prescribed by the scale and zero point can also have varied granularities, from per-tensor, per-group to per-channel.  
Furthermore, different optimization objectives could be used to determine scale and zero point.   
These extra parameters and procedures likely introduce additional variability into the scaling of PTQ of LLMs, making traditional \texttt{int} quantization more unpredictable than MX quantization.

With concrete examples, here we show that this is indeed the case.  
We create and calibrate \texttt{int} quantizers at varied precisions and granularities, denoted by \texttt{intP\_(chan|gG|tens)}.  
For example, \texttt{int4\_tens} represents a 4-bit per-tensor format, and \texttt{int3\_g32} a 3-bit per-group format with group size 32.  
We chose symmetric quantization scheme (with scale and no zero point) and calibrate by minimizing mean squared error (MSE) of quantization.  
Calibration data are 128 sequences taken from the training split.   

\begin{figure}[h!]
  \centering
  \includegraphics[scale=0.8]{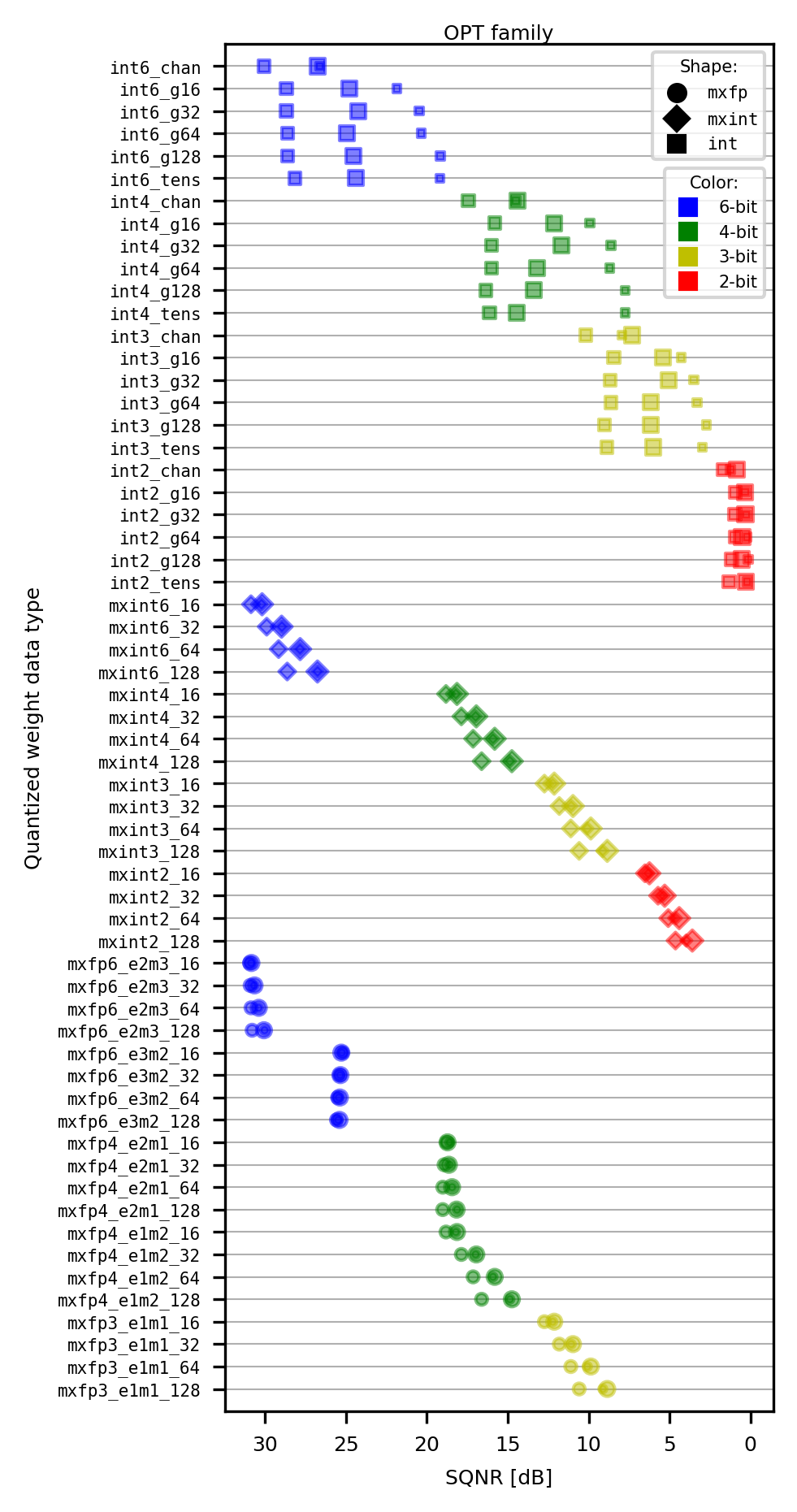}
  \caption{
    \textbf{SQNRs induced by traditional \texttt{int} versus MX quantizers for the smallest 3 models in the OPT family.}
    For notations of \texttt{int} formats and procedural details of calibration see the main text.  
    Numerical precision is color-coded and symbol sizes encode model capacity.  
  }
  \label{fig:int_vs_mx_sqnr}
\end{figure}

\begin{figure}[h!]
  \centering
  \includegraphics[scale=0.85]{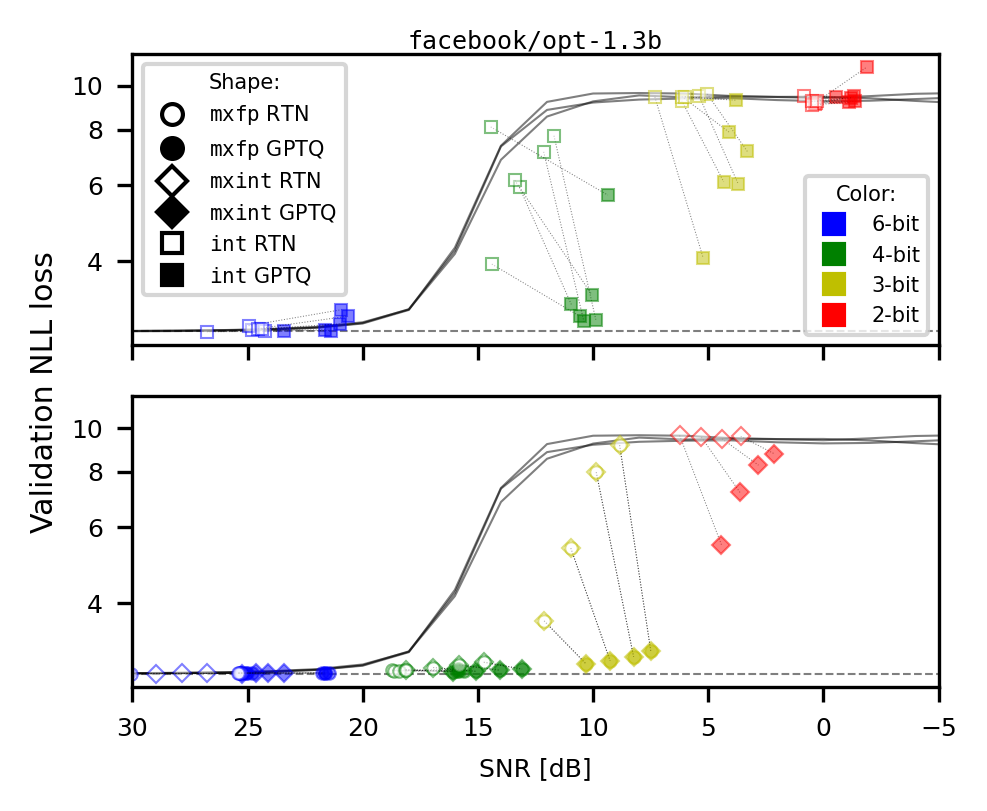}
  \caption{
    \textbf{Scaling of SQNRs and NLL losses before and after PTQ, for \texttt{int} versus MX data types.}
    Convention same as in Figure~\ref{fig:scaling_sqnr_nll_with_loss_landscape}.
    Data for \texttt{opt-1.3b} are shown, with \texttt{int} and MX formats separated in 2 panels. 
  }
  \label{fig:int_vs_mx_sqnr_nll}
\end{figure}

Not surprisingly, we find that SQNRs from \texttt{int} quantization are much more variable than those from MX quantization, and do not seem to scale monotonically with model size (Figure~\ref{fig:int_vs_mx_sqnr}). 
In addition, the changes to SQNRs and NLL losses as a consequence of GPTQ are much less predictable in the cases of \texttt{int} than MX data types (Figure~\ref{fig:int_vs_mx_sqnr_nll}).

\section{Interpretation of the importance of input features to the predictive model}
\label{sec:prediction_interpretation}

Beyond making accurate predictions of the difference in NLL loss between GPTQ and RTN, interpreting our predictive model can grant insight into the specific characteristics that make GPTQ most effective and the scenarios in which GPTQ should be employed.

\begin{figure}[h!]
  \centering
  \includegraphics[scale=0.85]{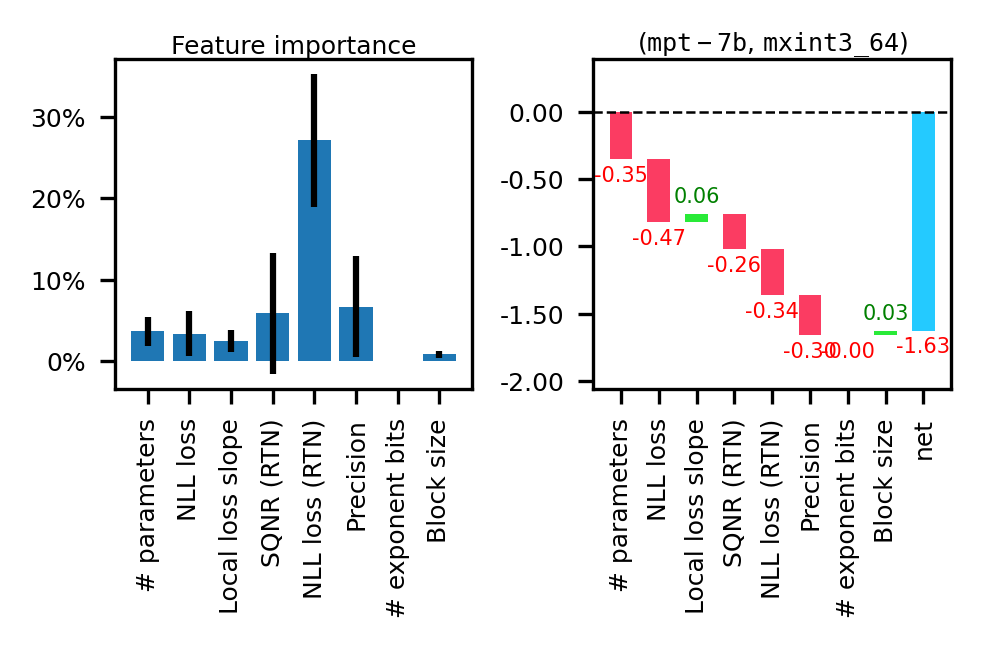}
  \caption{
    \textbf{Importance and interpretation of features used by our predictive model.}
    Mean and standard deviation of the importance score (Gini importance) for each input feature, calculated across all 120 trees in the random forest (left).  The predictive model's feature-specific decision-making process for quantizing \texttt{mosaicml/mpt-7b} to the \texttt{mxint3\_64} format (right).
  }
  \label{fig:prediction_interpretation}
\end{figure}

The Gini importance, also known as mean decrease in impurity, measures how much each feature contributes to reducing the Gini impurity in the dataset when making splits~\cite{gini}. As shown in (Figure~\ref{fig:prediction_interpretation}, left), our random forest regressor pays the most attention to the NLL loss of RTN, which can intuitively be explained by the understanding that GPTQ improves off of the baseline RTN quantization. Partial dependence graphs further reveal that the model pays more attention to the NLL loss of RTN at higher loss values, which is reasonable given that a higher starting NLL loss leaves greater room for GPTQ improvement. The number of parameters, the NLL loss of the original model, and the local loss slope are also considered by the predictive model because they describe the initial conditions of each LLM that differentiate their individual loss landscapes.

The quantization format accounts for three input features, namely precision, number of exponent bits, and block size. Of these features, precision has the largest influence on model prediction, which agrees with our findings that the largest variation in NLL loss between formats is driven by the number of bits (Figure~\ref{fig:gptq_sqnr_nll}, right). Note that the information gained from the quantization format is likely also embedded in the SQNR of RTN due to the strong correlation between SQNR and data format shown in (Figure~\ref{fig:rtn_sqnr_nll}, left), explaining why SQNR of RTN is also an important model feature.

The waterfall plot in (Figure~\ref{fig:prediction_interpretation}, right), highlights one example of how each input feature contributes to the random forest's prediction of the effect of GPTQ in quantizing the \texttt{mosaicml/mpt-7b} model to the \texttt{mxint3\_64} format.

\section{Cost of loss landscape feature computation}

\begin{figure}[h!]
  \centering
  \includegraphics[scale=0.85]{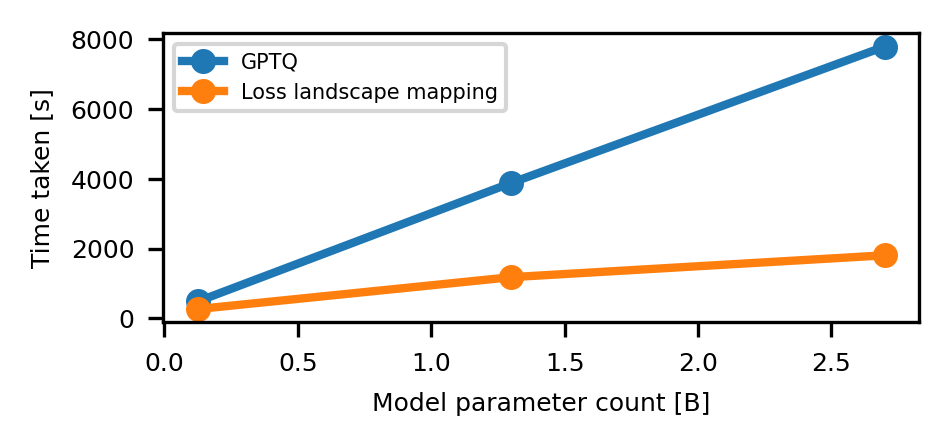}
  \caption{
   \textbf{Computational cost of GPTQ versus loss landscape mapping.} 
   We show data measured from runs of 3 models from the OPT family on a single A100 GPU, where time needed for loss landscape mapping is measured on 3 random weight perturbations.  
  }
  \label{fig:time}
\end{figure}

Our predictive model does not rely on features requiring second-order information, only empirical loss evaluation at critical points in the parameter space. 
Thus, only a few forward passes are needed to compute the input features to carry out a prediction, making the extraction of predictive features inexpensive.  
In Figure~\ref{fig:time}, we measure wall-clock time of feature extraction and compare it to conducting GPTQ optimization. We find that the overhead of running GPTQ is significantly more than measuring the step-wise loss landscape of 3 random weight perturbations, with the difference in overhead scaling with the model size. 
In practice, we only need loss landscape information local to the SNR of RTN, which could further reduce the amount of computation needed.
It is much more economical to use the predictive model based on scaling, than to actually compute GPTQ.

\section{Generalization across different datasets}
\label{datasets}
To verify the generalizability of our findings to other datasets, we repeated our experiments on the LAMBADA dataset~\citep{lambada} and a subset of the C4 dataset~\citep{c4}.

\begin{figure}[h!]
  \centering
  \includegraphics[scale=0.7]{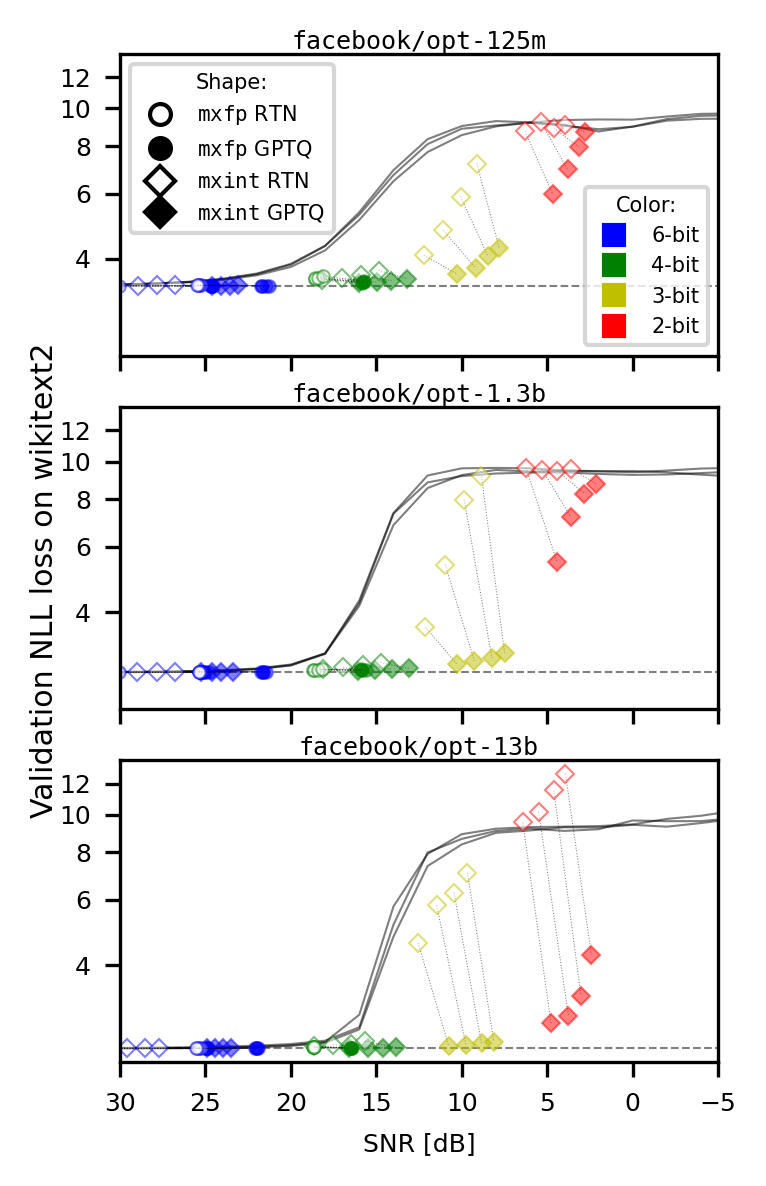}
  \includegraphics[scale=0.7]{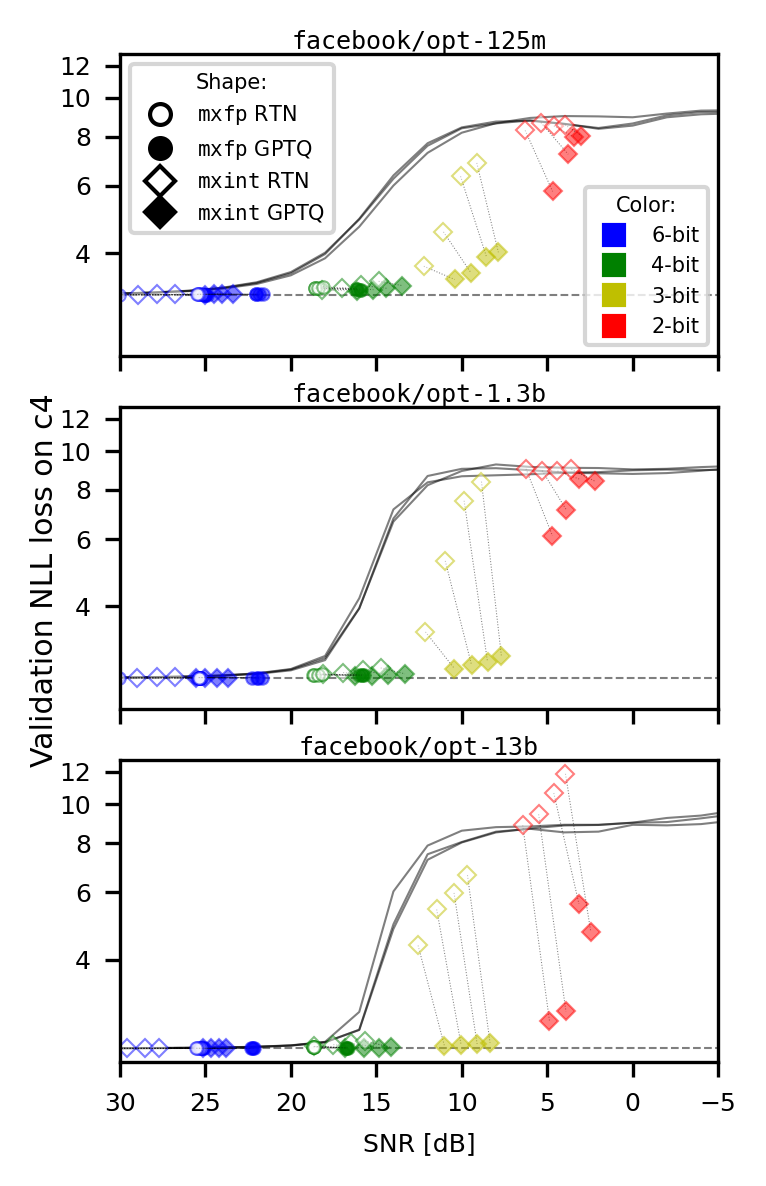}
  \includegraphics[scale=0.7]{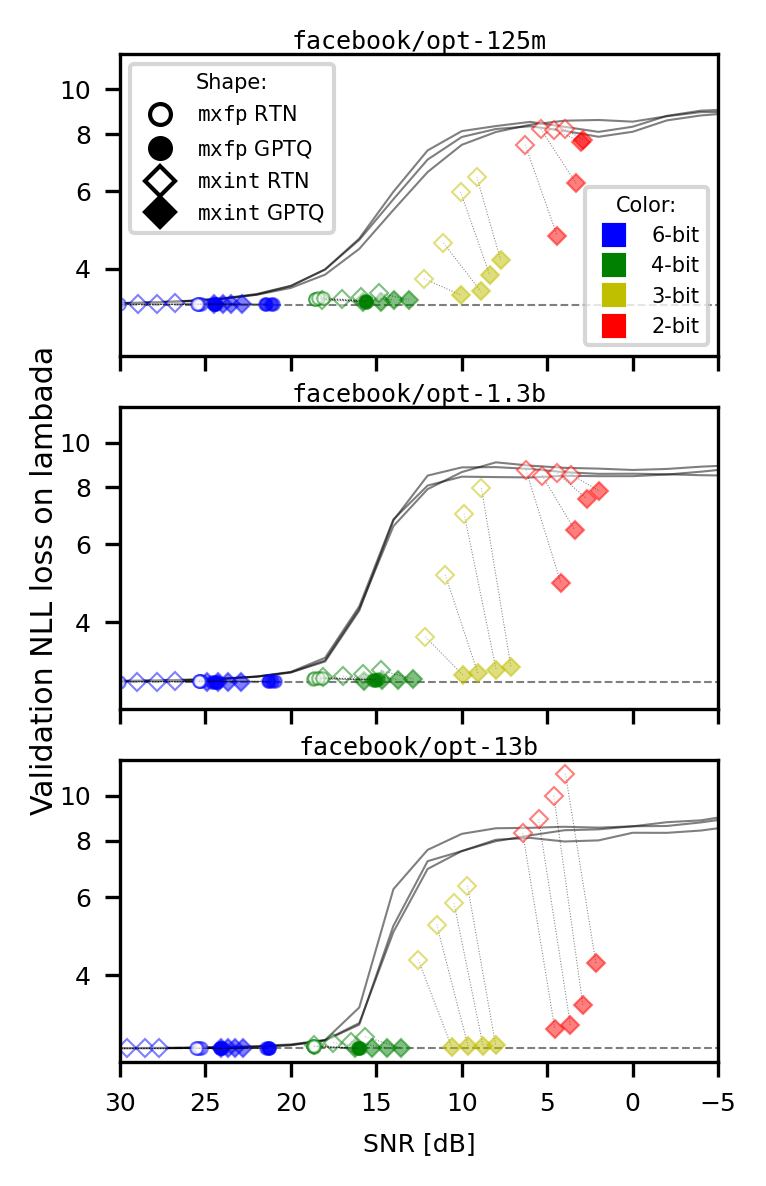}
  \caption{
    \textbf{Scaling of SQNRs and NLL losses before and after PTQ on different datasets}. Left: Wikitext2, middle: C4, right: LAMBADA.
  }
  \label{fig:nll-dataset}
\end{figure}

The scaling behavior of the SQNR and NLL losses displays similar trends across these datasets. In Figure~\ref{fig:nll-dataset}, we plotted the loss before and after applying GPTQ for three models in the OPT family. The pattern of the local loss landscape and the effect of GPTQ on different formats are consistent across all datasets.

\begin{figure}[h!]
  \centering
  \includegraphics[scale=0.65]{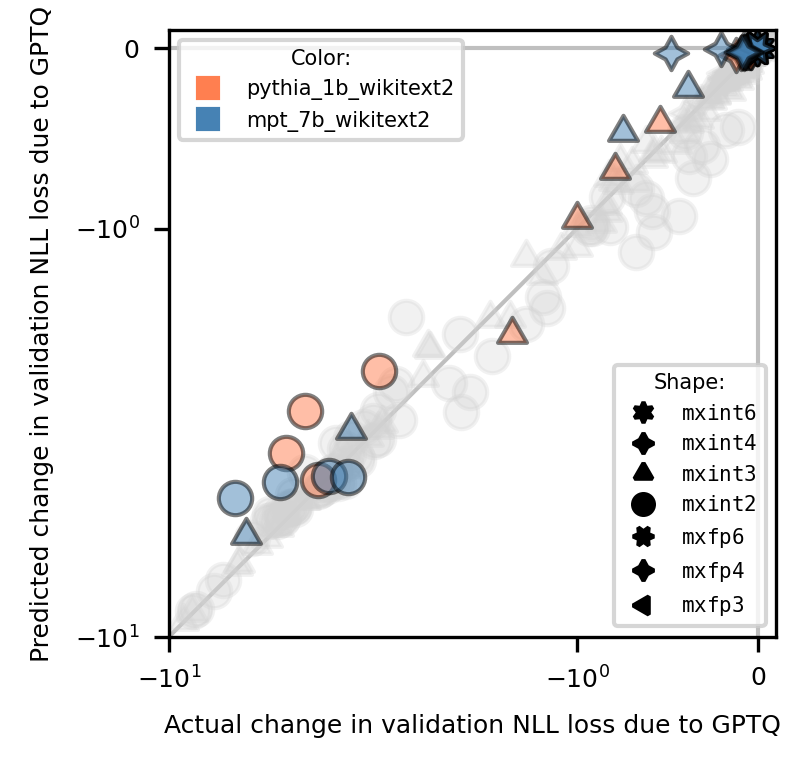}
  \includegraphics[scale=0.65]{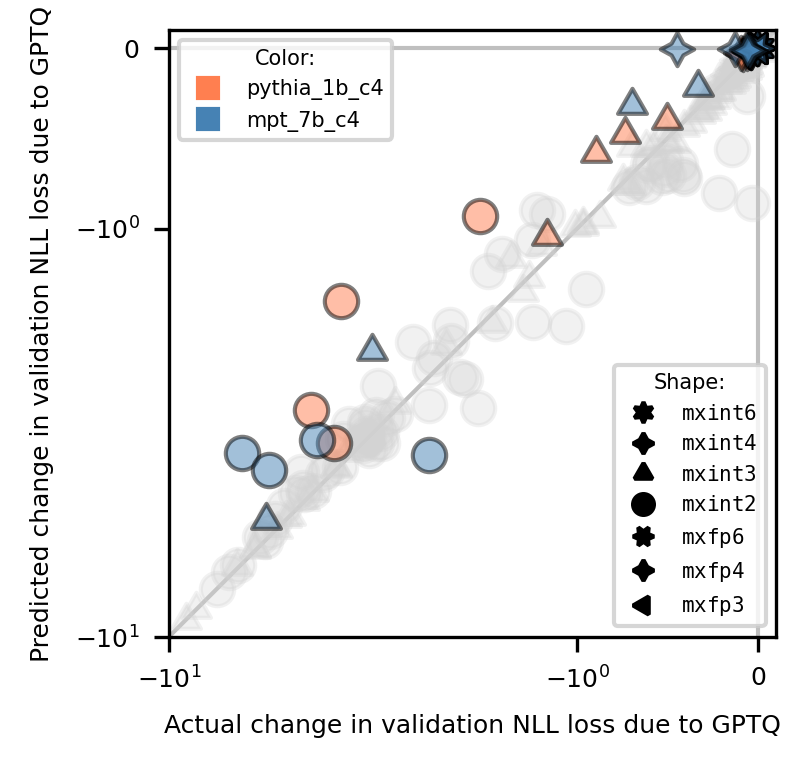}
  \includegraphics[scale=0.65]{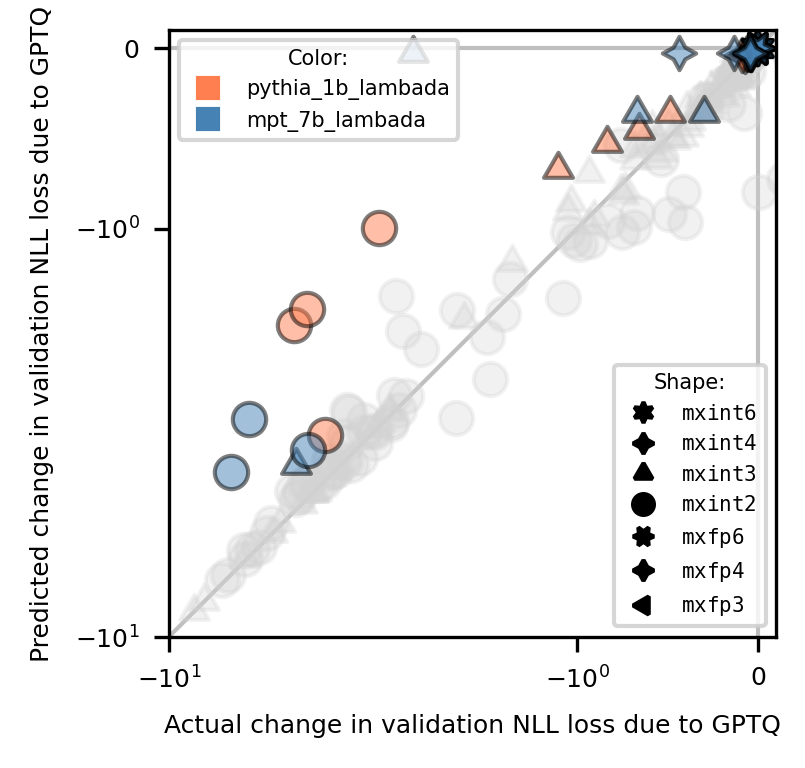}
  \caption{
    \textbf{Predictive models fitted on different datasets}. Left: Wikitext2, middle: C4, right: LAMBADA.
  }
  \label{fig:prediction-dataset}
\end{figure}

We further fitted predictive models on the new datasets. For each dataset, we trained a random forest regression predictor on all language models from the five model families and tested on two unseen language models. Figure~\ref{fig:prediction-dataset} shows that, similar to Wikitext2, the effect of GPTQ measured on C4 and LAMBADA can be modeled and predicted reasonably well. This indicates that capability of the random forest regression predictor can be generalized to other datasets.

\begin{figure}[h!]
  \centering
  \includegraphics[scale=0.8]{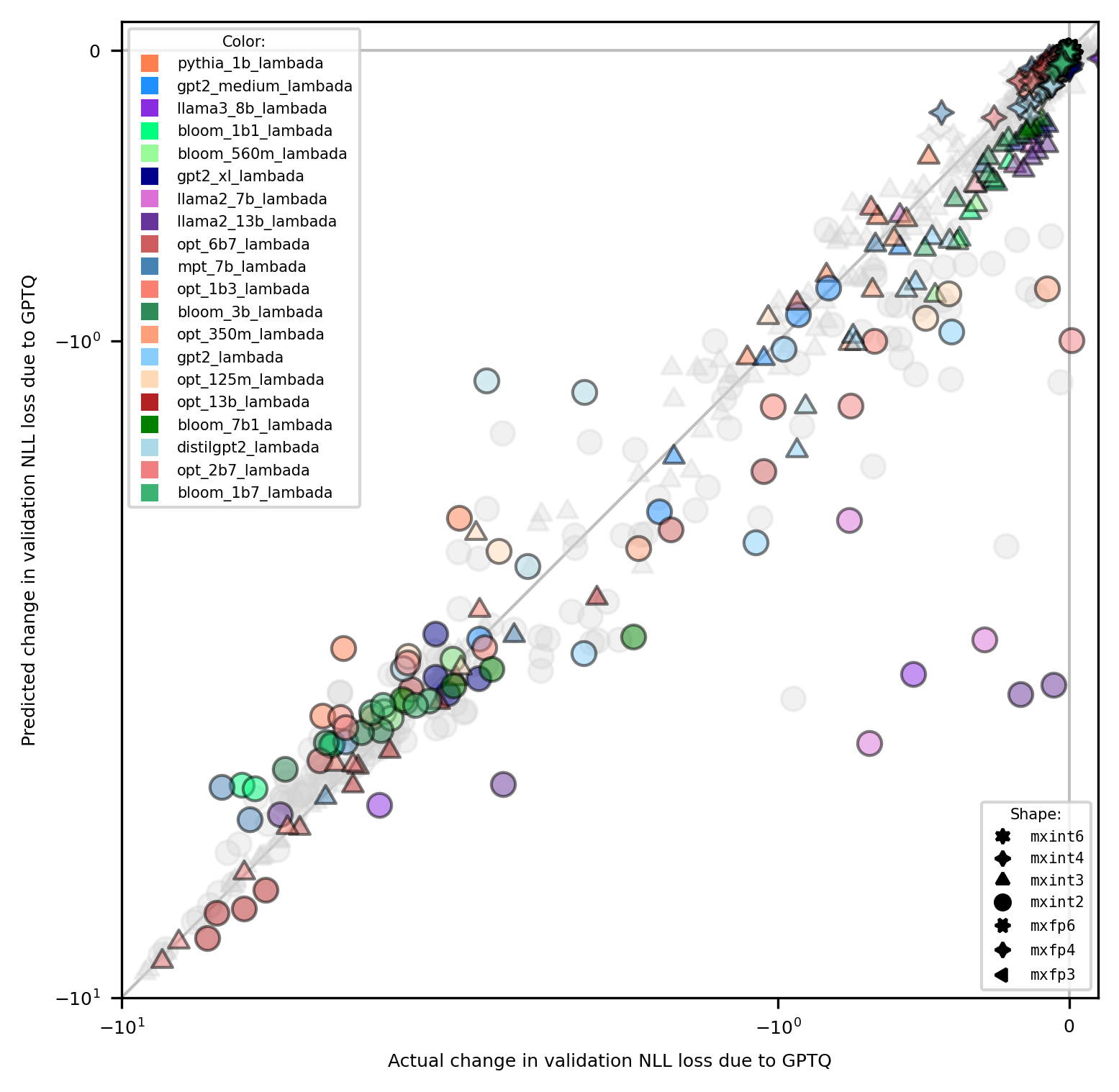}
  \caption{
    \textbf{Predictive model for cross-dataset prediction}. The predictive model is trained on data for all language models evaluated on Wikitext2 and C4 dataset (grey points) and predicted for the same models on LAMBADA dataset (colored points).
  }
  \label{fig:prediction-crossdataset}
\end{figure}

In addition, we verified cross-dataset prediction of the fitted regression model. We trained the regression model on all languages models evaluated on Wikitext2 and C4 dataset and tested its prediction for the same set of language models evaluated on the LAMBADA dataset. Figure ~\ref{fig:prediction-crossdataset} shows that the regression model is able to predict the effect of GPTQ accurately most of the time on an unseen dataset. 
\clearpage

\end{document}